\documentclass[lettersize,journal]{IEEEtran}
\usepackage{amsmath,amsfonts}
\usepackage{algorithmic}
\usepackage{algorithm}
\usepackage{array}
\usepackage[caption=false,font=normalsize,labelfont=sf,textfont=sf]{subfig}
\usepackage{textcomp}
\usepackage{stfloats}
\usepackage{url}
\usepackage{verbatim}
\usepackage{graphicx}
\usepackage{cite}
\usepackage{caption}
\usepackage{booktabs}
\usepackage{multirow}
\usepackage{multicol}
\usepackage{subfig}
\usepackage{float}
\usepackage{color}

\usepackage[dvipsnames,svgnames,x11names]{xcolor}

\usepackage[utf8]{inputenc}
\usepackage{amsmath}
\usepackage{amsthm}
\usepackage{booktabs}
\usepackage{algorithm}
\usepackage{algorithmic}
\usepackage{adjustbox}
\usepackage{hyperref}
\begin{document}

\title{Large Scale Time-Series Representation Learning via Simultaneous Low- and High-Frequency Feature Bootstrapping}

\author{Vandan Gorade,
        Azad Singh,~\IEEEmembership{Student Member, IEEE},
        Deepak Mishra,~\IEEEmembership{Member, IEEE}\\

\thanks{Vanddan Gorade and Azad Singh have contributed equally.}%
\thanks{
Vandan Gorade was with the Indian Institute of Technology Jodhpur, Jodhpur 342037, India. He is now with CVMI Lab, Jio Institute, Navi Mumbai 410206, India.
(e-mail: vangorade@gmail.com). Azad Singh and Deepak Mishra are with Computer Science and Engineering Department, Indian Institute of Technology Jodhpur, Karwar, 342037 India (e-mail: singh.63@iitj.ac.in, dmishra@iitj.ac.in).}
}

\markboth{IEEE Transactions on Neural Networks and Learning Systems, November~2023}%
{Gorade \MakeLowercase{\textit{et al.}}: Large Scale Time-Series Representation Learning via Simultaneous Low- and High-Frequency Feature Bootstrapping}

\maketitle

\begin{abstract}
Learning representations from unlabeled time series data is a challenging problem. Most existing self-supervised and unsupervised approaches in the time-series domain fall short in capturing low- and high-frequency features at the same time. As a result, the generalization ability of the learned representations remains limited. Furthermore, some of these methods employ large-scale models like transformers or rely on computationally expensive techniques such as contrastive learning. To tackle these problems, we propose a non-contrastive self-supervised learning approach that efficiently captures low- and high-frequency features in a cost-effective manner. The proposed framework comprises a siamese configuration of a deep neural network with two weight-sharing branches which are followed by low- and high-frequency feature extraction modules. The two branches of the proposed network allow bootstrapping of the latent representation by taking two different augmented views of raw time series data as input. The augmented views are created by applying random transformations sampled from a single set of augmentations. The low- and high-frequency feature extraction modules of the proposed network contain a combination of multilayer perceptron (MLP) and temporal convolutional network (TCN) heads respectively, which capture the temporal dependencies from the raw input data at various scales due to the varying receptive fields. To demonstrate 
the robustness of our model, we performed extensive experiments and ablation studies on five real-world time-series datasets. Our method achieves state-of-art performance on all the considered datasets.
\end{abstract}


\section{Introduction}
\IEEEPARstart{T}{ime}-series data such as electroencephalogram (EEG), electrocardiogram (ECG), or data collected from wearable devices are extensively used in various fields, particularly health care. With the advancement of technology, the volume of captured time-series data is rapidly expanding; thus, computer-aided algorithms are needed to assist physicians in making accurate and timely diagnoses.
Deep learning models~\cite{faust2018deep,8614252,8437249,zhang2022tn} outperform traditional time series analysis methods such as applied statistical and machine learning. However, training data-hungry deep learning models with a limited amount of annotated samples is difficult. Unlike natural images, understanding and annotating time-series data is challenging and expensive and often requires a domain expert~\cite{Ching142760,chang2021comprehensive}. Further, supervised deep learning models with small annotated datasets often learn the task-specific representations which have limited generalizability.

Self-supervised learning (SSL) has emerged as a popular choice due to its ability to learn
meaningful generalized representations from the underlying structure of unlabeled data itself~\cite{oord2018representation, moco,simclr,li2022video}. SSL encourages unsupervised pre-training followed by data-efficient supervised fine-tuning with a limited set of annotated data thus, provides a way to leverage abundant unlabeled time-series data generated in daily clinical practices. Recently it gained a lot of attention in various domains like computer vision, natural language processing, graphs, speech, etc., as it achieved near state-of-the-art supervised learning performance. SSL based upon the nature of pre-training, is mainly divided into three paradigms.

\textit{(I)} \textit{Predictive}: Pretext objectives relevant to the domain and the input data are designed first in the SSL pre-training part, which is then solved by a model to learn generic representations. The learned representations are then subsequently applied to the different downstream tasks, such as object classification and detection. For example,~\cite{jigsaw} proposed to learn representation by solving a jigsaw puzzle,~\cite{exemplecnn} proposed Exempler-CNN,~\cite{rotpred} prediction of the rotated angle of the input image,~ \cite{colorization} proposed image colorization pretext task. Although solving the pretext task enables the model to learn meaningful representations however identifying and designing the right pre-text task is itself a challenge. Also, pretext tasks are not generalizable across different domains and tasks; for example, if the pretext task is to predict the rotation angle of an image, it may deviate the model from learning other features like color and distortions~\cite{cpc}.

\textit{(II)} \textit{Generative}: In this, the pretext task is designed to reconstruct the original input while learning the meaningful latent representation. In~\cite{ contextautoenc}, authors proposed a context autoencoder, which is trained to fill in the missing piece in an image. In~\cite{brainsplit}, Zhang et al.  proposed split-brain autoencoders to learn representation by predicting a subset of color channels from the rest of the channels. These types of tasks are often hard to train and inefficient since the model parameters frequently oscillate and rarely converge~\cite{liu2021self}.

\textit{(III)} \textit{Discriminative}: In contrast to generative models, discriminative approaches are led by contrastive learning~\cite{simclr, moco,cpc}. Contrastive learning is an instance-level discriminative approach, where the basic idea is to learn transformation invariant representations by pushing the augmented samples from the same input image (positive pairs) towards each other in the embedding space, while those belonging to different input images (negative pairs) are pushed apart. Recent advances in contrastive learning have shown prominent results. Recently non-contrastive approaches~\cite{byol,simsiam} have also emerged to mitigate the inherent challenges of contrastive learning by relying only on positive pairs to learn effective representations.

Existing discriminative SSL approaches are developed explicitly for image or text data since there are multiple augmentation strategies available to create positive and negative pairs. Recently, some works based on contrastive learning~\cite{sslecg,selfhar,seqclr} for EEG, ECG, and other time series data have been proposed. However, augmentations from image and text modality are not readily transferable to time-series data due to its inherent heterogeneity in magnitude, frequency, and time~\cite{franceschi2020unsupervised,iwana2021empirical}. Time series data, in general, need domain and problem-specific augmentations, as augmentations like color distortion, superpixel, and cartoonization may not work for time series data. Further, most existing methods cannot capture low- and high-frequency features simultaneously, which are essential for learning effective representations~\cite{franceschi2020unsupervised, wavelets}.

To overcome the aforementioned challenges, we propose a non-contrastive SSL approach for large-scale time-series representation learning  via simultaneous bootstrapping of low- and high-frequency input features. The approach is motivated by BYOL~\cite{byol}, which circumvents the issues of contrastive learning, \textit{i.g.} costly training, by relaxing the requirement of negative pairs. To simultaneously capture the low- and high-frequency features, we employed low-and high-frequency feature extraction modules using multilayer perceptron (MLP) and temporal convolutional network (TCN)~\cite{tcn} heads, respectively. The idea of introducing low- and high-frequency feature extraction modules in a non-contrastive self-supervised learning framework is driven by the nature of time-series data. Therefore, it results in an efficient and generalized approach to learning time-series representation and works across different datasets.
Our contributions are summarized as follows.

\begin{itemize}
\item We propose a novel non-contrastive SSL method for analyzing the time-series data, which is simple yet effective and can work without a large pool of labeled data.

\item Our method can capture low- and high-frequency features at the same time. It uses MLP and TCN heads which capture temporal dependencies at various scales in a complementary manner to learn effective representations.

\item We perform extensive experiments using a wide range of real-world time-series datasets. Experimental results show that the learned representations are effective for downstream tasks under both linear evaluation and semi-supervised settings of SSL.
\end{itemize}

\section{Related Works}

\subsection{Representation learning}
Bengio et al.~\cite{RL} reported that a good representation is one that disentangles the underlying factors of variation and constructs a space that is discriminative for downstream tasks. It is based on the idea of better network convergence by adding (unsupervised) pre-trained vectors that encode mutual information from the multiple interacting underlying features. This work led various approaches in domains like Computer vision, Natural language processing (NLP), Speech, and Time-Series ~\cite{kingma2013auto,ashfahani2020devdan,miotto2016deep,deng2013recent,shewalkar2019performance,yan2021zeronas,li2022video}. With the recent advances in deep learning; researchers can now capture the semantics of time series data from the healthcare domain. Med2Vec~\cite{choi2016multi} and Wave2Vec~\cite{yuan2019wave2vec} are two popular approaches that capture the interpretable representations from the healthcare data with a temporal aspect. Inspired by word embeddings, Med2Vec treats medical codes (medication, diagnosis) as discrete words. It learns representations for concise medical codes by leveraging skip-gram embeddings~\cite{mikolov2013distributed} along with longitudinal representations of patient visits. Wave2Vec, on the other hand, focuses on continuous waveform biosignals (ECG, EMG). The authors used a fixed-size sliding window to segment the waveform into fragments and preprocess it using wavelet transform before training a unified model. The model learns temporal relationships between signal fragments.

\subsection{Self-supervised Learning} Self-supervised learning is a paradigm in which the model is first trained in an unsupervised way and followed by supervised fine-tuning or transfer learning using small labeled data. Generally speaking, self-supervised learning based on the nature of pre-training divides into three main types, predictive, generative, and contrastive. A wide range of predictive and generative tasks have been proposed to learn representation like rotation prediction~\cite{rotpred}, jigsaw puzzle~\cite{jigsaw}, context encoders~\cite{contextautoenc}, split-brain autoencoders~\cite{brainsplit} and colorization~\cite{colorization}. However, predictive and generative pretext tasks are application specific and not generalizable. In contrast to hand-crafted pretext tasks, contrastive learning learns invariant and robust feature representation using augmentations.
The leading approaches for self-supervised learning based upon contrastive paradigm are Momentum-contrast (MoCo)~\cite{moco} and Simple contrastive-learning (SimCLR)~\cite{simclr}, which learn representation by maximizing the similarity of different views from the same sample while simultaneously minimizing the similarity among views from different input samples. More recently proposed approaches like Bootstrap your own latent (BYOL)~\cite{byol}, SimSiam~\cite{simsiam}, and Barlow Twins~\cite{zbontar2021barlow} learn representation with only positive pairs. 

\subsection{Self-supervised Learning for Time-Series}
SSL for Time-Series has recently received attention from researchers. A wide range of pretext tasks has been explored to learn good time-series representation. For example, SSL-ECG ~\cite{sslecg} predict transformations similar to rotation prediction~\cite{rotpred} pretext task. Similarly~\cite{selfhar} proposed transformation prediction task for human activity recognition. Inspired by the success of contrastive learning in other domains,~\cite{cpc} proposed contrastive predictive coding, which learned context latent representations that encode the underlying information shared between different parts of the input, by predicting the discrete future representations in the latent space. 
It showed great advances in various speech recognition tasks. Also,~\cite{seqclr} extended SimCLR model~\cite{simclr} to EEG data. More recently~\cite{tstcc} proposed multitask contrastive learning approach named TS-TCC, which captures temporal and contextual information from time series. These approaches are either not generalizable or need labeled data. Also, existing approaches failed to simultaneously capture low- and high-frequency features, which are essential characteristics of time series.~\cite{franceschi2020unsupervised}. The proposed approach addresses all the above-mentioned issues effectively.

\begin{figure*}[!ht]
    \centering
    \includegraphics[width=140mm,scale=1.8]{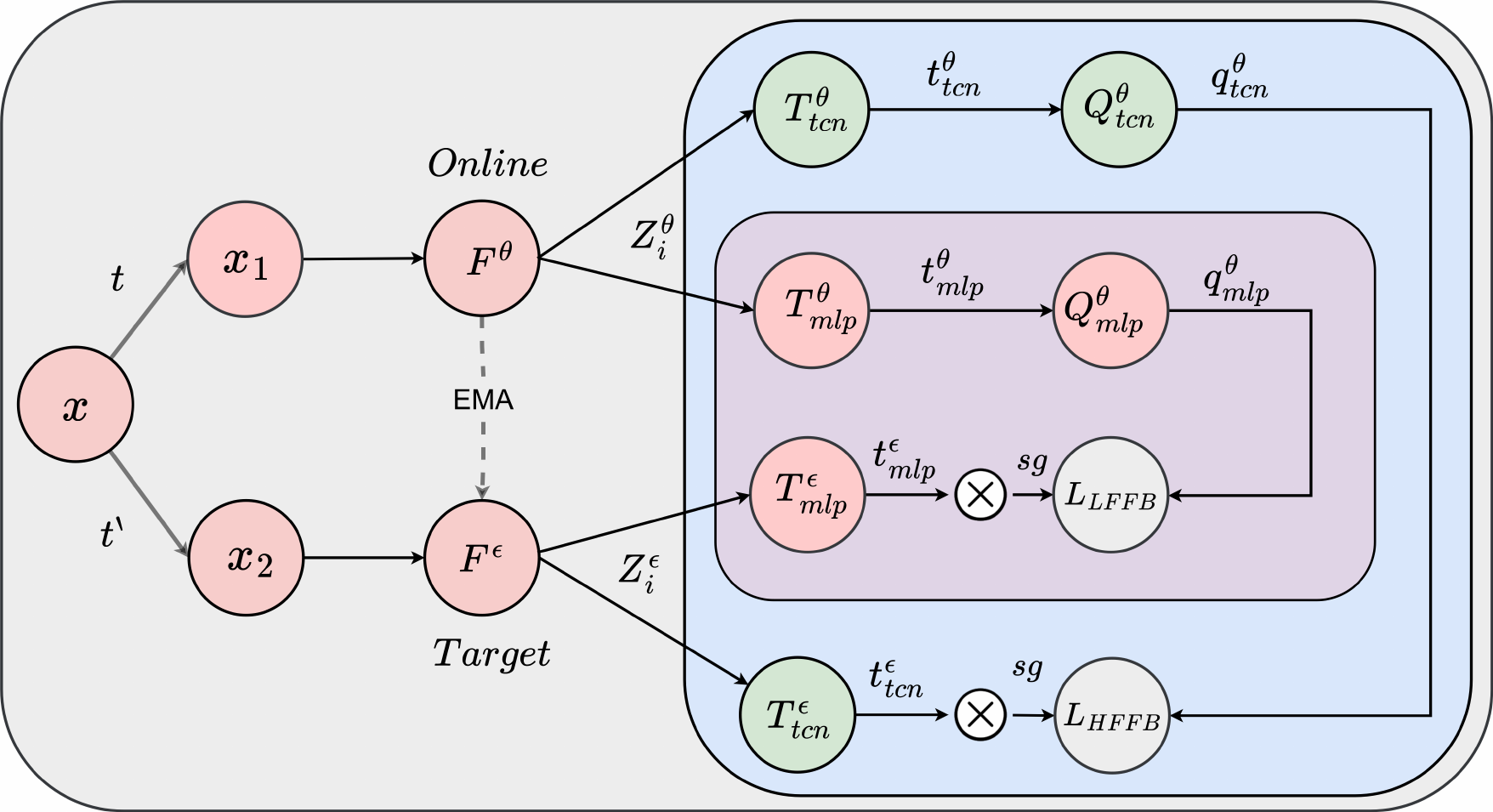} 
    \caption{Given an input sample x, model creates two augmented versions $x_1 \sim t$ and $x_2 \sim t^{'}$, where $t$ and $t^{'}$ are sampled from a set $T$ of augmentations. $x_1$ and $x_2$ pass through the backbone encoders parameterized by $\theta$ and $\epsilon$ which give $Z^\theta$ and $Z^\epsilon$ as output.
     Further, $Z^\theta$ and $Z^\epsilon$ are processed by the LHFFB module to learn  complementary representations. LHFFB constitutes of LFFB and HFFB modules to capture low- and high-frequency features from latent representations respectively.
     EMA denotes exponential moving average and $sg$ denotes stop gradient.}
    \label{fig1}
    
\end{figure*}

\section{Proposed Method}

This section describes the details of our proposed method. As shown in Fig.~\ref{fig1} we first generate two different augmented versions from input time-series data. These augmented versions pass through weight-sharing backbone encoders named online $F^\theta$ and target $F^\epsilon$ networks, which are three-block convolution networks, followed by low-frequency and high-frequency feature bootstrapping modules. The proposed method has a large kernel size in backbone CNN encoders which enables them to capture rich representations of the input time series data. Representations learned by the backbone encoders have both low and high frequency components which are further decomposed by the low- and high-frequency bootstrapping modules.    

The low-frequency feature bootstrapping (LFFB) module using MLP heads is responsible for capturing features from longer time intervals i.e. having low-frequency. The low-frequency regions indicate a general trend that constitutes the global semantic features. Particularly the LFFB module is composed of three multilayer perceptron (MLP) heads: two for the online branch, $T^{\theta}_{mlp}$ and $Q^{\theta}_{mlp}$, and one for the target branch, $T^{\epsilon}_{mlp}$.

Similar to LFFB,  the high-frequency feature bootstrapping module (HFFB) encompasses two temporal convolution network (TCN)~\cite{tcn} heads named as $T^{\theta}_{tcn}$ and $Q^{\theta}_{tcn}$ in the online branch while one in target branch named as $T^{\epsilon}_{tcn}$. HFFB is responsible for learning time-varying features in higher-frequency (shorter time period) windows where the data contains discontinuities, ruptures, and singularities. 

TCN is a simple convolutional approach. It consists of dilated causal convolution operations, which make it an efficient method for the analysis of time series data. Similar to recurrent neural network (RNN), TCN can also map input sequences of any length to the same length output sequence. Further dilation and causal convolution in TCN offer large receptive fields, which makes it suitable to capture longer historical information with a faster learning speed with lower memory requirements~\cite{bai2018empirical}. As shown in ~\cite{zhang2021temporal,yan2020temporal,dai2022price} that due to dilated convolutions, TCN can capture higher frequency components efficiently, which have local information and strong randomness.

Both LFFB and HFFB modules capture complementary features, which results in making the model robust. The target network provides regression targets to train the online network while its own parameters $\epsilon$ update as an exponential moving average of the online parameters $\theta$. More precisely as in the equation below,

\begin{equation}
    \epsilon \leftarrow \tau\epsilon + (1 - \tau)\theta.
    \label{eq:1}
\end{equation}

We perform above mentioned update for each training step with a decay rate $\tau \in [0,1]$.
Each module is explained in detail in the following subsections (~\ref{dga},~\ref{sec:hffb},~\ref{sec:lffb},~\ref{sec:lhfb}).

\subsection{Domain-guided Augmentations}
\label{dga}
Data augmentation is a crucial component of SSL, where the goal is to learn viewpoint-invariant representations. By creating diverse but related versions of the input data while maintaining semantic meaning, data augmentations enable the SSL framework to capture representations that remain invariant to the augmented versions~\cite{moco,simclr,simsiam,byol}. In the context of SSL for time series data, we generate two augmented views, $x_1$ and $x_2$,  by applying data augmentations, including jittering, permutations, and rotations, where Jittering is simply an addition of random noise, typically drawn from a Gaussian distribution to the time series data at each time point while keeping the time steps constant. Permutation in the time series data refers to shuffling the different time slices of the data points while preserving their values. Rotation in the context of multivariate time series data can be applied by utilizing an element-wise random rotation matrix $R$ with a defined angle, while in the case of univariate time series data, rotation can be achieved by applying circular shifting to the data, which involves shifting the entire time series data circularly. Mathematically, given a univariate time series $X$ = $\{x_1,x_2,x_3, ……, x_t\}$, circular shifting by $k$ time steps can be achieved by moving the last $k$ elements of $x$ to the front, creating a circular permutation. In case of multivariate time series data  $X'$ = $X * R$, i.e. $X'$ = $\{x_1*R, x_2*R, ……, x_t*R \}$. These augmentations are chosen according to the specific characteristics of time series data to enhance the diversity in $x_1$ and $x_2$. As reported in Table~\ref{table:dataset} out of 5, two datasets are multivariate (HAR and IMU ), while three (Sleep-EDF, Epilepsy, and ECG ) are univariate datasets. 

For the input to the $F^\theta$, we introduce random variations to the signal through jittering and split the signal into a random number of segments (up to a maximum of M), and randomly shuffle them to apply permutation. Subsequently, a rotation of $30^\circ$ is applied to the augmented views. Similarly, the same set of augmentations is performed on the input to the $F^\epsilon$, but with a rotation angle of $45^\circ$. The choice of different rotation angles for online and target network branches is inspired by the BYOL architecture of SSL, where the idea is to learn generalized representations which are invariant to different transformations. By rotating the input data for both online and target networks with different angles, the proposed approach introduces a range of transformations that the network needs to nullify to learn effective representations. The choice of rotation angles, specifically $30^\circ$ for the online network and $45^\circ$ for the target network, has been empirically determined based on the preliminary experiments. These values aim to strike a balance between introducing enough variability in the input and maintaining the coherence and integrity of the time series data.

\subsection{HFFB Module}
\label{sec:hffb}
This module is responsible for capturing high-frequency features from the input time series signal, which has shorter time intervals with discontinuities, singularities, and ruptures.
Specifically as shown in Fig~\ref{fig1} $x_1$ and $x_2$ pass through $F^\theta$ and $F^\epsilon$ respectively to extract high-dimensional latent representation $Z^\theta$ and $Z^\epsilon$. Particularly $Z^\theta = [z_{0}^{\theta}, z_{1}^{\theta}...z_{n}^{\theta}]$ and $Z^\epsilon = [z_{0}^{\epsilon}, z_{1}^{\epsilon}...z_{n}^{\epsilon}]$, which can be denoted as:

\begin{equation}
    z_{i}^{n} = F^{n}_{i}(x_{i}^{n})
    \label{eq:2}
\end{equation}

Where, $i$ denotes number of time-stamps and $n \in (\theta, \epsilon)$. $F^{n}$ consists of $N$ convolutional layer with a large kernel followed by batchnorm, ReLu activation, and a pooling layer.

Now, given representations $Z^\theta$ and $Z^\epsilon$ for each $x_1$ and $x_2$  projection heads $T^{\theta}_{tcn}$ and $T^{\epsilon}_{tcn}$, capture high-frequency  representations, $t^{\theta}_{tcn}$ and $t^{\epsilon}_{tcn}$ as depicted in equation~\eqref{eq:3}.

\begin{equation}
    t_{i}^{n} =N \times [TCN(BNorm(ReLU(z_{i}^{n})), K, D)]
    \label{eq:3}
\end{equation}

In equation~\ref{eq:3}, $K$ and $D$ parameters denote kernel and dilation rate, while the other parameters are the same as in equation~\eqref{eq:2}. We keep $K$ small for TCN while the dilation rate $D$ changes hierarchically. 

This enables the model to capture high-frequency features as the dilation rate increases the receptive field of the TCN hierarchically. Further $t^{\theta}_{tcn}$ are then passed through $Q^{\theta}_{tcn}$ which is responsible for predicting the embeddings from $T^{\epsilon}_{tcn}$. 
To update the parameters $\theta$ of $F^\theta$, mean squared error is applied between $q^{\theta}_{tcn}$ and $t^{\epsilon}_{tcn}$, which are the representations learned by $Q^{\theta}_{tcn}$ and  $T^{\epsilon}_{tcn}$ respectively, as shown in the equation~\eqref{eq:4}
\begin{equation}
    L_{HFFB} =  \|\tilde{q}^{\theta}_{tcn} - \tilde{t}^{\epsilon}_{tcn}\|^2_2 
    \label{eq:4}
\end{equation}
\begin{equation} 
    L_{HFFB} = 2 - 2 \cdot \frac{\langle	 \tilde{q}^{\theta}_{tcn}, \tilde{t}^{\epsilon}_{tcn}\rangle	}{\|\tilde{q}^{\theta}_{tcn}\|.\|\tilde{t}^{\epsilon}_{tcn}\|}
    \label{eq:5}
\end{equation}
where $\tilde{q}^{\theta}_{tcn}$ and $\tilde{t}^{\epsilon}_{tcn}$ are the $l_{2}$ normalized version of $q^{\theta}_{tcn}$ and $t^{\epsilon}_{tcn}$ respectively.

\subsection{LFFB Module}
\label{sec:lffb}
This module is responsible for capturing low-frequency features from the latent representation produced by the backbone encoder using MLP head.
As can be seen from equation~\eqref{eq:2} projection heads  $T^{\theta}_{mlp}$ and $T^{\epsilon}_{mlp}$ employs both representations respectively to generate latent low dimensional lower frequency representation $m^{\theta}_{mlp}$ and $m^{\epsilon}_{mlp}$ as shown in equation \eqref{eq:6}.

\begin{equation}
    m^{n}_{i} = N \times MLP(z^{n}_{i})
    \label{eq:6}
\end{equation}
where $MLP$ consists of one linear layer followed by Batchnorm, ReLU, and another linear layer. 
Further $m^{\theta}_{mlp}$ is passed through $Q^{\theta}_{mlp}$ which is same as in equation~\eqref{eq:6} and responsible for predicting the embeddings generated from $T^{\epsilon}_{mlp}$. After that, to update the parameters $\theta$ of $F^\theta$ mean squared error is calculated between $q^{\theta}_{mlp}$ and $m^{\epsilon}_{mlp}$, where $q^{\theta}_{mlp}$ is the representations learned by $Q^{\theta}_{mlp}$.  The equation ~\eqref{eq:7} shows the mean squared error between $l_{2}$ normalized $q^{\theta}_{mlp}$ and $m^{\epsilon}_{mlp}$.

\begin{equation}
    \mathcal{L}_{LFFB} =  \|\tilde{q}^{\theta}_{mlp} - \tilde{m}^{\epsilon}_{mlp}\|^2_2 
    \label{eq:7}
\end{equation}
\begin{equation}
    \mathcal{L}_{LFFB} = 2 - 2 \cdot \frac{\langle \tilde{q}^{\theta}_{mlp}, \tilde{m}^{\epsilon}_{mlp}\rangle	}{\|\tilde{q}^{\theta}_{mlp}\|.\|\tilde{m}^{\epsilon}_{mlp}\|}
    \label{eq:8}
\end{equation}

where $\tilde{q}^{\theta}_{mlp}$ and $\tilde{m}^{\epsilon}_{mlp}$ are the $l_{2}$ normalized version of $q^{\theta}_{mlp}$ and $m^{\epsilon}_{mlp}$ respectively.

\subsection{Low- and High-Frequency Feature Bootstrapping Module}
\label{sec:lhfb}
This module combines both LFFB and HFFB to capture complementary features. The motivation of high- and low-frequency modules is to capture both low- and high-frequency features. For example, in the wake stage and stage 1 of sleep, signals are of low frequency (up to 4Hz), and as we move into stage 2 sleep, the body goes into a state of deep relaxation, and we start to see bursts of activity known as sleep spindles. A sleep spindle is a rapid burst of higher frequency brain waves~\cite{sleepstage}. Stage 3 and stage 4 of sleep are often referred to as deep sleep or slow-wave sleep because these stages are characterized by low-frequency (up to 4 Hz). Similarly, in ECG signals, it is important to detect R-peaks. These are typically the highest peaks in an ECG signal. Similarly, in human activity recognition (HAR), sitting and running correspond to low- and high-frequency signals.

In TCN, the convolution operation is applied to the input time series data using fixed-size convolutional kernels. TCN consists of multiple layers with dilated causal convolutions, where each layer captures information at different temporal scales. Mathematically, the convolution operation in TCN can be represented as:
\begin{equation}
(f * w)(t) = \sum_{i=1}^{N} f(t-d \cdot i) \cdot w_i
\label{eq:tcn}
\end{equation} 
where $f(t)$ is the input time series data, $w_i$ represents the learnable convolutional filter weights, $*$ denotes the convolution operation, $t$ is the time parameter, and $d$ is the dilation rate. The convolution operation is performed by sliding the convolutional kernel over the input sequence with a fixed step size determined by the dilation rate. This allows the network to capture information from a larger receptive field without increasing the number of parameters or the computational complexity. 

Continuous wavelet transform (CWT) is a signal processing technique that also provides a hierarchical representation of time series data in a different way than TCN. In CWT, the convolution operation is performed between the input time series signal and the wavelet function at different scales and translations. Mathematically, the CWT of a signal $f(t)$ with respect to a wavelet function $g(t)$ can be written as:
\begin{equation}
(f * g)(t) = \frac{1}{\sqrt{d}}\int_{-\infty}^{\infty} f(\tau)g\left(\frac{\tau - t}{d}\right) d\tau
\label{eq:cwt}
\end{equation}
where $\tau$ represents the translation factor, and $d$ is the dilation or scale factor. The wavelet function $g$ is continuous both in time and frequency domains. The convolution is performed over the entire range of $\tau$ to capture the similarities between the input signal and the wavelet function at different scales and translations ($t$). 

To relate TCN and CWT, we can rewrite equation~\ref{eq:tcn} similar to equation~\ref{eq:cwt} as:
\begin{equation}
(f * w)(t) = \sum_{i=1}^{N} f(\tau) \cdot w_i \quad \text{where} \quad \tau = t - d \cdot i
\end{equation}
As we see that the wavelet function $g$ in equation~\ref{eq:cwt} is dependent on the scale parameter $d$ and the difference between $\tau$ and $t$. Furthermore, $f(\tau)$ in continuous convolution equation~\eqref{eq:cwt} corresponds to $f(t-d \cdot i)$ in equation~\eqref{eq:tcn} in the sampled form. Similarly, $g\left(\frac{\tau - t}{d}\right)$ in the continuous convolution equation can be interpreted as the weights $w_i$ in the discrete convolution form. 

Both TCN and CWT involve convolution operations between the input signal and certain functions (kernels in TCN and wavelet functions in CWT). However, CWT involves convolving the input signal with the wavelet at different scales and translations. This operation is computationally expensive, especially for long-range signals or large datasets. Also, CWT relies on selecting appropriate scales and translations for the wavelet. Choosing the right parameters is challenging and requires prior knowledge about the signal characteristics. Furthermore, CWT typically involves a separate feature extraction step where relevant features are extracted from the transformed signal. This requires domain expertise and manual feature engineering. Also, CWT's time-frequency representation provides information about the signal at specific time points and frequency bands but may not capture long-term dependencies or contextual information over a larger time window. In contrast, TCN operates on fixed-size convolutional kernels, making it computationally efficient, especially for long signals or large datasets. TCN learns to extract features directly from the input data in an end-to-end manner. It eliminates the need for explicit feature engineering or a separate feature extraction step, as CWT requires. This makes TCN more flexible and adaptable to different signal classification tasks without relying on prior domain knowledge. Also, TCN is designed to capture long-range dependencies in the input signal effectively. Using dilated convolutions, TCN can exponentially increase the receptive field as the network depth increases. This enables the network to model and learn dependencies across a larger time window, allowing it to capture long-term temporal patterns in the data. Additionally, TCN architecture can be easily scaled up or down to handle different input sequence lengths. With longer input sequences, TCN can increase its depth or adjust the kernel size to capture adequate historical information without encountering gradient explosion or vanishing gradients. This scalability makes TCN suitable for tasks involving variable-length time series data. We employed various experimental settings to validate the behavior of TCN and reported the results in Table~\ref{tab:table2} and~\ref{tab:table3}. 

Similar observations can be made for MLP by considering dilation $d$ = 1 with several 1D kernels, which enable it to capture lower frequency features. Thus, MLP and TCN module work in tandem to simultaneously extract low- and high-frequency learned representation while utilizing BYOL architecture. The complete model enables learning complex features from raw signals without external supervision. Therefore, this module is responsible for capturing both types of features from data. The low- and high-frequency feature bootstrapping is as follows:

\begin{equation}
L_{LHFFB} = \lambda \times L_{LFFB} + (1-\lambda) \times L_{HFFB}
\end{equation}
where $\lambda$ is a fixed scalar hyperparameter that denotes each loss's relative weight.

\section{Experimental Setup}
\subsection{Descriptions of Datasets}
In our experiments, we used five datasets, Human Activity Recognition (HAR)~\cite{anguita2013public}, Sleep-EDF~\cite{goldberger2000physiobank}, Epilepsy~\cite{andrzejak2001indications}, IMU sensor dataset \cite{IMU} and Mobile Health dataset (ECG MEDH)~\cite{Dua:2019}. The HAR dataset accumulates six human activities, which include walking, sitting, standing, walking$\_$upstairs, and walking$\_$downstairs from 30 individuals in the 19-48 years age group. The dataset was collected using an embedded accelerometer and gyroscope of the smartphone with a sampling rate of 50 Hz. The sleep-EDF dataset contains whole-night PolySomnoGraphic sleep recordings divided into five classes according to sleep stages. In our experiments, we use a single EEG channel with a sampling rate of 100 Hz. IMU sensor data is collected while driving a small mobile robot where the task is to predict which one of the nine-floor types (carpet, tiles, concrete) the robot is on using sensor data such as acceleration and velocity. The mobile health dataset has ECG measurements of body motion and vital signs from ten volunteers while performing 12 physical activities. Table~\ref{table:dataset} provides more details regarding datasets.
The important aspect of these datasets is that they contain both low- and high-frequency components, where the high-frequency components correspond to sudden bursts or rapid fluctuations in the signal, while the low-frequency components exhibit limited fluctuations and interruptions. It is worth noting that we did not explicitly pre-process the data to extract the low- and high-frequency features.
Further, to ensure a fair comparison with the baselines, we adopted the pre-processing steps employed by TS-TCC. These steps mainly involved scaling and normalization of the data. To maintain consistency, we utilized the pre-processing scripts provided by TS-TCC, which are publicly accessible in their GitHub repository [\href{https://github.com/emadeldeen24/TS-TCC}{link}]. This approach established a standardized and comparable experimental setup across the different datasets. Further, in our deep learning-based SSL approach, we employed a combination of CNN, TCN, and MLP networks for automatic feature extraction from raw time-series data. These neural network architectures are designed to capture different aspects of the data and extract meaningful representations without requiring extensive manual feature engineering. Combining these network architectures allows the proposed approach to automatically learn and extract discriminative features from raw time-series data. This reduces the reliance on manual feature engineering, simplifies the preprocessing steps, and potentially improves the model's ability to capture complex patterns and representations in the data. 

\begin{table}[ht]
    \centering
    \caption{Description of datasets used in our experiments}
    \label{table:dataset}
    \begin{adjustbox}{width=0.47\textwidth}
    \begin{tabular}{ccccc}
        \toprule
         \multirow{1}{*}{Datasets} & \multicolumn{1}{c}{Samples} &
         \multicolumn{1}{c}{Length} &
         \multicolumn{1}{c}{Channels} &
         \multicolumn{1}{c}{Classes} \\
         
        \midrule
        
        HAR &  10299 & 128 & 9 & 6 \\
        Sleep-EDF & 11500 & 3000 & 1 & 5\\
        Epilepsy &  11500 & 178 & 1 & 2\\
        IMU Sensor Data & 4578 & 128 & 9 & 10\\
        ECG MEDH & 1021225 & 23 & 1 & 13\\

        \bottomrule
    \end{tabular}
    \end{adjustbox}
\end{table}

\subsection{Implementation Details}
All datasets are split into 60\%, 20\%, and 20\% for training, validation, and testing. We follow the training strategy of~\cite{tstcc}. We train our model for 40 epochs on HAR and Elipesy for both pretraining and downstream training. We used batch sizes from 128 to 32 depending on the size of the dataset. For SleepEDF and ECG MEDH datasets, we pretrain our model for 20 and 50 epochs, respectively, and downstream training is done for 40 and 50 epochs, respectively. We set a batch size of 150 for both of them. For the IMU sensor dataset, pretraining and downstream training is done for 100 epochs each. For all datasets, we used Adam optimizer with a learning rate of 3e-4, weight decay of 3e-4, $\beta_1$ = 0.9, and $\beta_2$ = 0.99. We set $\lambda$ = 0.51 for all datasets. The dropout rate for all datasets is 0.35. We used two layers of TCN for both projection and prediction heads, and for each layer, we used kernel size and dilation rate of 50 and $2_i$ where $i$ is a hidden dimension for TCN. Following~\cite{byol,simsiam}, each MLP head has two linear layers. To ensure a fair comparison, we adopted the hyperparameters specified in the baseline paper TS-TCC as a starting point. However, for any hyperparameters not mentioned in the baseline paper, we determined them through empirical experimentation.
\begin{table*}[!ht]
    \fontsize{9}{10}\selectfont
    \centering
    \caption{ \fontsize{9}{10}\selectfont Performance of our model on HAR, SleepEDF, and Epilepsy dataset for linear evaluation under Accuracy and MacroF1 metrics. It is worth noting that we have trained all the baselines from scratch under the same hyperparameters setting. It can be observed that the proposed SSL pre-training approach consistently outperforms all the defined baselines by a considerable margin. }
    \begin{adjustbox}{width=1\textwidth}
    \begin{tabular}{c c c c c c c}
        \toprule
         \multirow{1}{*}{Methods} &
         \multicolumn{2}{c}{HAR} &  \multicolumn{2}{c}{Sleep-EDF} &
         \multicolumn{2}{c}{Epilepsy} \\
         \cmidrule{2-7} 
        & ACC & MF1 & ACC & MF1 & ACC & MF1 \\ 
        \toprule

        Random Initialization & 77.77$\pm$4.31\% & 77.12$\pm$4.54\% & 37.84$\pm$7.48\% & 28.72$\pm$9.12\% & 92.52$\pm$2.57\% & 87.61$\pm$5.42\% \\
        Supervised  & 92.47$\pm$2.31\% & 92.27$\pm$2.10\% & \textbf{83.41$\pm$1.44\%} & 74.78$\pm$0.86\% & 97.10$\pm$0.44\% & 96.64$\pm$0.53\%\\
        \midrule
        CPC \cite{cpc} & 83.85$\pm$1.51\% &  83.27$\pm$1.66\% & 82.82$\pm$1.68\% & 73.94$\pm$1.75\% & 96.61$\pm$0.43\% & 94.44$\pm$0.69\%\\
        
        SimCLR \cite{simclr} & 80.97$\pm$2.46\%  & 80.19$\pm$2.64\% & 78.91$\pm$3.11\% & 68.60$\pm$2.71\% & 96.05$\pm$0.34\% & 93.53$\pm$0.63\% \\
        
        SimSiam \cite{simsiam} & 91.74$\pm$0.27\% & 91.68$\pm$0.31\% & 80.10$\pm$2.15\% & 
        72.34$\pm$0.60\% & 
        93.84$\pm$0.52\% & 
        94.04$\pm$0.15\%\\
        
        TS-TCC \cite{tstcc} & 91.13$\pm$0.54\% & 90.98$\pm$0.33\% & 82.90$\pm$0.41\% &  73.49$\pm$1.24\% & 97.33$\pm$0.24\% & 96.14$\pm$0.42\%\\

        \midrule
        Ours (TCN + MLP) & 93.74$\pm$0.30\% & \textbf{93.28$\pm$0.25}\% & 82.84$\pm$0.65\% & 
        \textbf{76.54$\pm$0.90}\% & \textbf{98.14$\pm$0.22}\% & 
        \textbf{97.54$\pm$0.35}\%\\
        
        Ours (TCN + No MLP Head) & 92.72$\pm$0.29\% & 92.56$\pm$0.34\% & 80.24$\pm$1.15\% & 
        72.64$\pm$1.95\% & 
        96.34$\pm$ 0.12\% & 
        96.14$\pm$ 0.10\%\\
        
        Ours (MLP + No TCN Head) & 92.34$\pm$0.47\% & 92.28$\pm$0.51\% & 79.64$\pm$1.95\% & 
        72.74$\pm$1.20\% & 
        94.44$\pm$0.12\% & 
        94.30$\pm$0.55\%\\

        Ours (TCN \& MLP + Conv Backbone )& 
        \textbf{93.76$\pm$0.34}\% &
        91.11$\pm$0.39\% & 
        80.23$\pm$0.58\% &
        71.54$\pm$0.24\% & 
        98.13$\pm$0.11\% &
        98.12$\pm$0.13\% \\

        \bottomrule
    \end{tabular}
    \end{adjustbox}
    \label{tab:table2}
    
\end{table*}

\begin{table*}[ht]
    \fontsize{9}{10}\selectfont
    \centering
    \caption{ \fontsize{9}{10}\selectfont This Table is the extension the Table ~\ref{tab:table2}, which shows the performance of our model on IMU sensor, and ECG MEDH datasets for Linear evaluation under Accuracy and MacroF1 metrics. We observe from the table that our proposed SSL pre-training approach consistently outperforms all baselines for both datasets.}
    \begin{adjustbox}{width=0.75\textwidth}
    \begin{tabular}{c c c c c}
        \toprule
         \multirow{1}{*}{Methods} &
         \multicolumn{2}{c}{IMU Sensor Data} &  \multicolumn{2}{c}{ECG MEDH} \\
         \cmidrule{2-5} 
        & ACC & MF1 & ACC & MF1  \\ 
        \toprule

        Random Initialization & 
        20.77$\pm$4.31\% & 
        15.12$\pm$4.54\% & 
        68.69$\pm$2.45\% & 
        54.57$\pm$3.75\%\\
        Supervised  & 
        74.57$\pm$2.31\% & 
        73.10$\pm$2.10\% & 
        87.11$\pm$0.24\% &
        77.99$\pm$0.14\% \\
        \midrule
        CPC \cite{cpc} & 28.39$\pm$1.03\% &  21.49$\pm$0.46\% & 72.01$\pm$0.47\% & 60.55$\pm$0.63\% \\
        
        SimCLR \cite{simclr} & 66.34$\pm$2.51\%  & 55.61$\pm$3.1\% & 86.32$\pm$1.03\% & 76.29$\pm$0.96\%  \\
        Simsiam \cite{simsiam} & 
        72.64$\pm$0.19\% & 
        72.15$\pm$0.67\% & 
        85.40$\pm$0.24\% &
        75.85$\pm$0.53\%\\
        
        TS-TCC \cite{tstcc} &
        37.97$\pm$2.34\% & 
        24.58$\pm$3.29\% & 
        84.57$\pm$0.34\% &
        61.60$\pm$0.65\%\\

        \midrule
        Ours (TCN + MLP) & 
        \textbf{75.84$\pm$1.50}\% & 
        73.28$\pm$0.25\% & 
        \textbf{87.81$\pm$0.25\%} &
        78.20$\pm$0.25\%\\
        
        Ours (TCN + No MLP Head) & 
        73.58$\pm$1.39\% & 
        73.06$\pm$2.42\% & 
        86.84$\pm$0.15\% &
        76.90$\pm$0.35\%\\
        
        Ours (MLP + No TCN Head) & 
        72.94$\pm$0.89\% & 
        72.66$\pm$1.33\% & 
        85.90$\pm$0.80\% &
        76.75$\pm$0.23\%\\

        Ours (TCN \& MLP + Conv Backbone) & 
        75.68$\pm$0.38\% & 
        \textbf{73.48$\pm$0.46}\% & 
        87.79$\pm$0.41\% &
        \textbf{78.24$\pm$0.63}\%\\
                
        \bottomrule
    \end{tabular}
    \end{adjustbox}
    \label{tab:table3}
    
\end{table*}

\subsection{Baselines}
 We compare our proposed approach with baselines, including supervised (random initialization, supervised), contrastive ( Contrastive predictive coding (CPC)~\cite{cpc}, SimCLR~\cite{simclr}), and non-contrastive (SimSiam~\cite{simsiam}, BYOL~\cite{byol}) SSL approaches. We also include the current top-performing method TS-TCC~\cite{tstcc}. In random initialization, we train a linear classifier on top of a randomly initialized backbone encoder, whereas in supervised, we train both the backbone encoder and linear classifier with labeled data. To make a fair comparison, we keep the hyperparameter setting identical across all the experiments. Further, we did not train BYOL exclusively, as our proposed method without a TCN head is equivalent to BYOL.

\section{Results and Discussion}
\subsection{Evaluation Protocols}
Following TS-TCC~\cite{tstcc}, to demonstrate the robustness and efficiency of our proposed method, we employed two different evaluation protocols: (1) Linear evaluation and (2) Semi-supervised evaluation. These evaluation protocols have been widely used in the field of SSL~\cite{moco,simclr,byol,simsiam,tstcc}. 

In the linear evaluation protocol, we trained a single linear layer classifier on top of the backbone CNN encoder. It is important to note that during this evaluation, the parameters of the backbone CNN encoder are kept frozen, ensuring that only the parameters of the linear layer are updated. This setup allows us to assess the discriminative power of the representations learned by the backbone CNN encoder during pretraining for their effectiveness in a downstream classification task. By isolating the linear layer for downstream training, we could measure the impact of the self-supervised pretraining on enhancing the model's ability to classify instances accurately. On the other hand, in the semi-supervised evaluation protocol, we fine-tuned the pretrained backbone CNN encoder along with the newly added linear layer classifier, using different subsets of randomly sampled data from the training set. Specifically, as reported in the paper, we utilized 1\%, 5\%, 10\%, and 50\% of the training samples to fine-tune the model. 
By fine-tuning the model on this limited labeled data while leveraging the knowledge gained from self-supervised pretraining, we aimed to evaluate the model's ability to generalize and make accurate predictions on unseen or partially labeled data. This evaluation protocol provides insights on the proposed method's performance under scenarios where labeled data is scarce or expensive to obtain. 

Further, to effectively measure the performance of our proposed method, we utilized two widely used evaluation metrics: accuracy and MF1 (Macro F1) score. Accuracy is a commonly used metric that quantifies the proportion of correctly classified instances in the evaluation dataset, providing an overall measure of classification performance. However, accuracy alone may not provide a comprehensive evaluation of the performance of the time-series models, particularly when dealing with imbalanced datasets. Time-series data often exhibits class imbalance, where certain classes have a larger number of instances compared to others. To address this, we also used the MF1 score, which calculates the F1 score for each class individually and then averages them, assigning equal importance to each class. The F1 score combines precision and recall into a single metric using the harmonic mean as: F1 = 2 * (precision * recall) / (precision + recall).

\subsection{Results}
We report the performance of the proposed model for linear evaluation in Table~\ref{tab:table2} and~\ref{tab:table3}, where standard deviation indicates results for three runs with different seeds. 
The model compared with different baselines include a fully-supervised method, CPC, SimCLR, SimSiam, and TS-TCC, and different variations of the proposed approach. These baselines represent state-of-the-art methods in the field of self-supervised learning for time-series data. The evaluation metrics used are the accuracy and MacroF1 (MF1) score, which provides insights into the classification performance of the models.

Comparing the results, it can be observed that the proposed SSL approach consistently outperforms all baselines in terms of both accuracy and MF1 score across all the considered datasets. This indicates the effectiveness and superiority of the proposed SSL pre-training in capturing relevant patterns and learning effective representations from the time-series data, which leads to improved classification performance. For instance, in the HAR dataset, the proposed approach achieves an accuracy of 93.74\%, while the best-performing SSL baseline achieves an accuracy of 91.74\%. This is a considerable improvement in classification accuracy by the proposed approach compared to the SSL baselines. Similarly, in terms of the MF1 score, the proposed SSL approach achieves a relatively higher score of 93.28\% as compared to the SSL baselines. It is worth noting that the proposed approach is also able to surpass the fully supervised method with a margin of more than 1\% in terms of both accuracy and MF1 score. Similarly, in the Sleep-EDF dataset, our approach achieves 76.54\% MF1 score, which is an average of 2.5\% higher than all the baselines, including fully supervised methods and TS-TCC as well. Further, in the Epilepsy dataset, the proposed approach achieves an accuracy of 98.14\% and an MF1 score of 97.54\%. Once again, the proposed approach outperforms all the baselines with a margin of more than 1\%, showcasing its ability to learn meaningful representations. 

Furthermore, the evaluation of the IMU Sensor Data and ECG MEDH datasets also demonstrates the superiority of the proposed SSL approach. In both datasets, our approach achieves higher accuracy and MF1 scores compared to the baselines, especially in comparison with TS-TCC, which is considered to be the state-of-the-art for certain other datasets. This suggests that the baselines struggle to capture the intricate patterns and relationships present in these specific types of time-series data. On the other hand, the proposed approach is able to capture the unique characteristics and patterns specific to these datasets, resulting in more accurate and informative representations. 

It is important to note that TS-TCC, which shows competitive performance in HAR, Sleep-EDF, and Epilepsy datasets,
incurs a higher computation cost compared to the proposed SSL approach. This is primarily due to a large number of model parameters, excessive memory requirements, and high computation demands associated with TS-TCC's architecture. TS-TCC utilizes a multi-task contrastive loss and employs transformers as the backbone encoder. While transformers have shown promising results in various domains, they typically require a large number of parameters to achieve their full potential. As a result, TS-TCC ends up with a large number of model parameters, which increases the computational complexity and memory requirements. On the other hand, the proposed SSL approach, which combines TCN and MLP architectures, offers a more efficient alternative. The TCN architecture, known for its parallelism and efficient processing of sequential data, provides effective feature extraction capabilities. This combination strikes a balance between model complexity and computational efficiency, resulting in a more practical and scalable solution. Therefore, in addition to outperforming the baselines in terms of performance metrics, the proposed SSL approach offers the advantage of lower computation cost also.

\begin{figure}
\begin{tabular}{cc}
  \includegraphics[width=40mm]{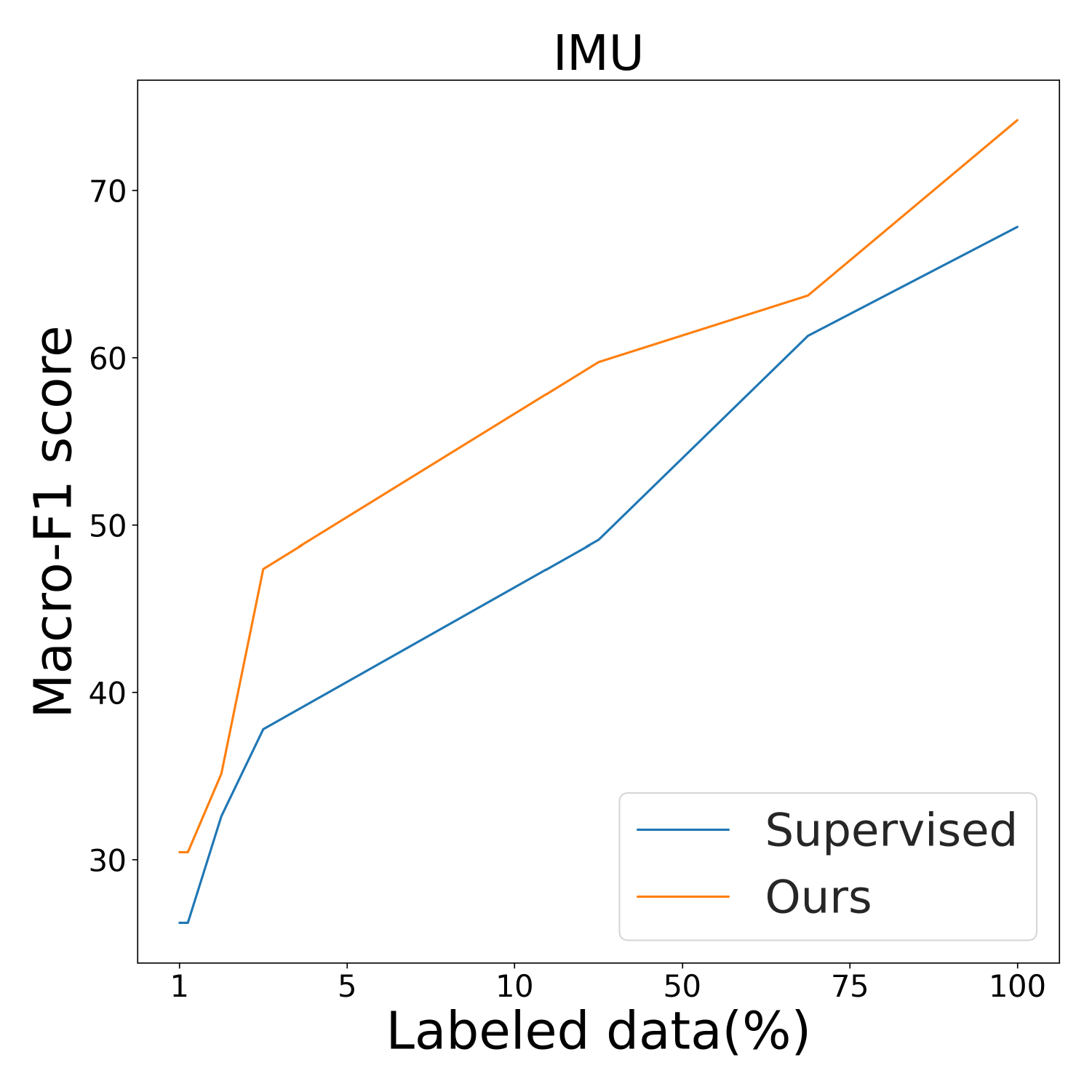} &   \includegraphics[width=40mm]{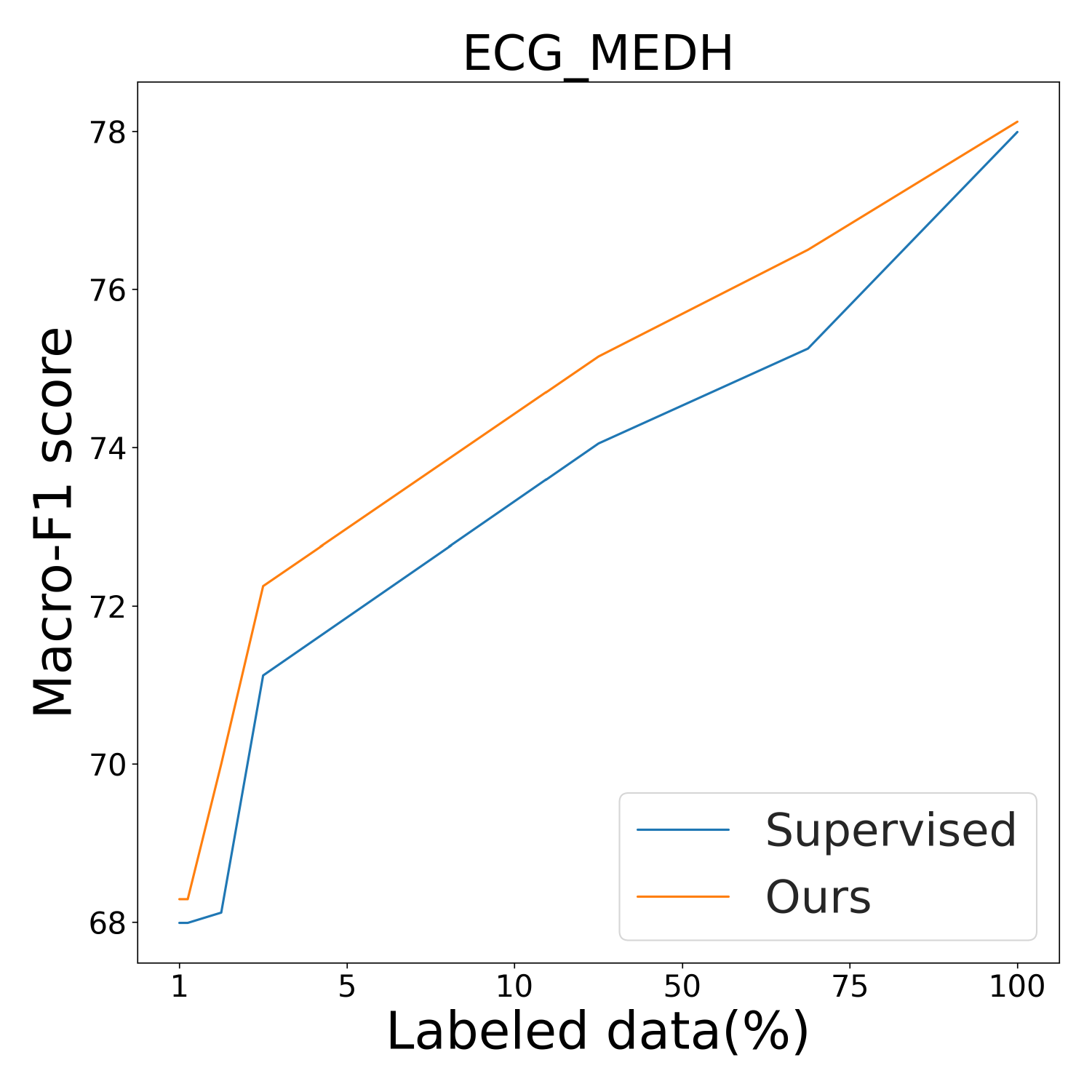} \\
 \includegraphics[width=40mm]{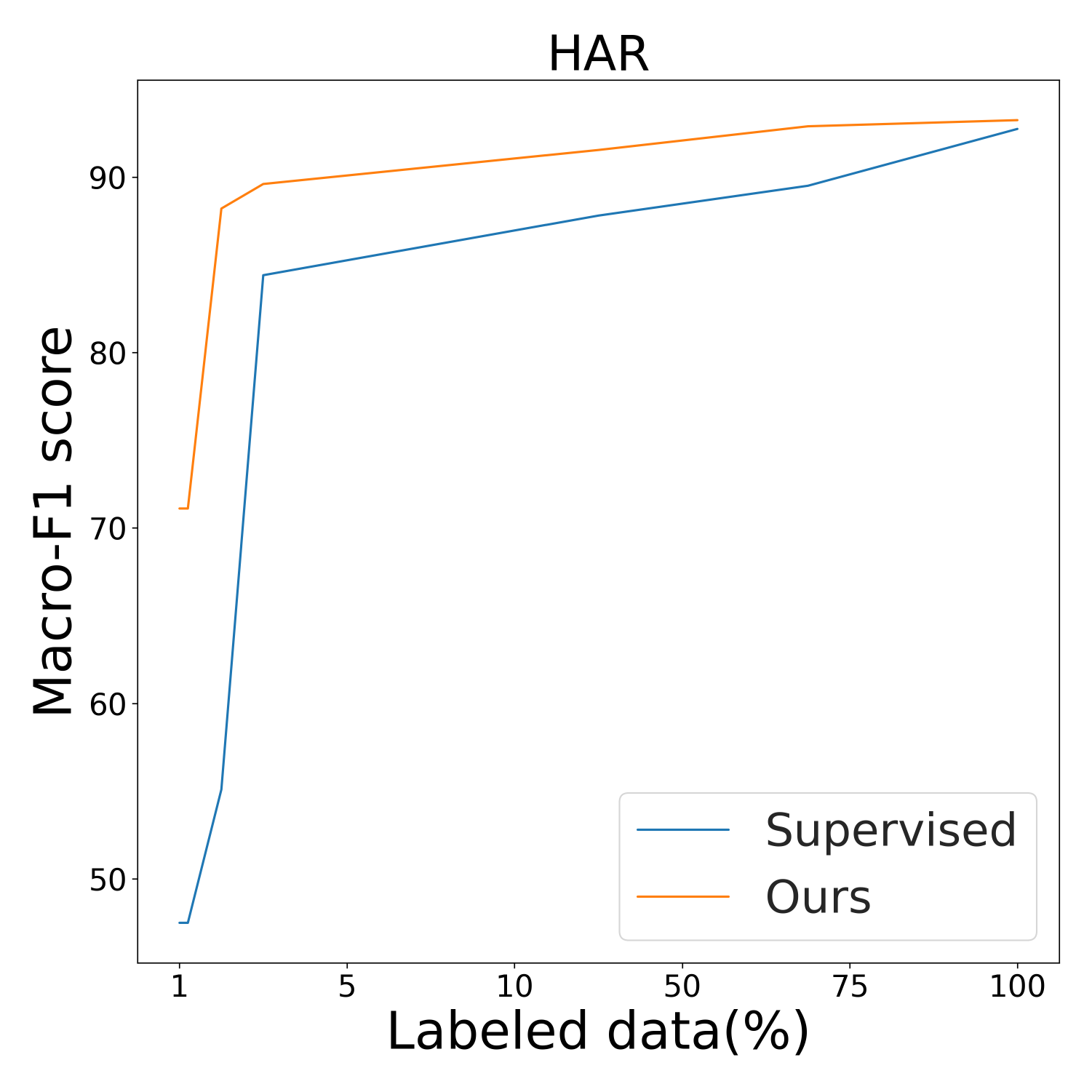} &   \includegraphics[width=40mm]{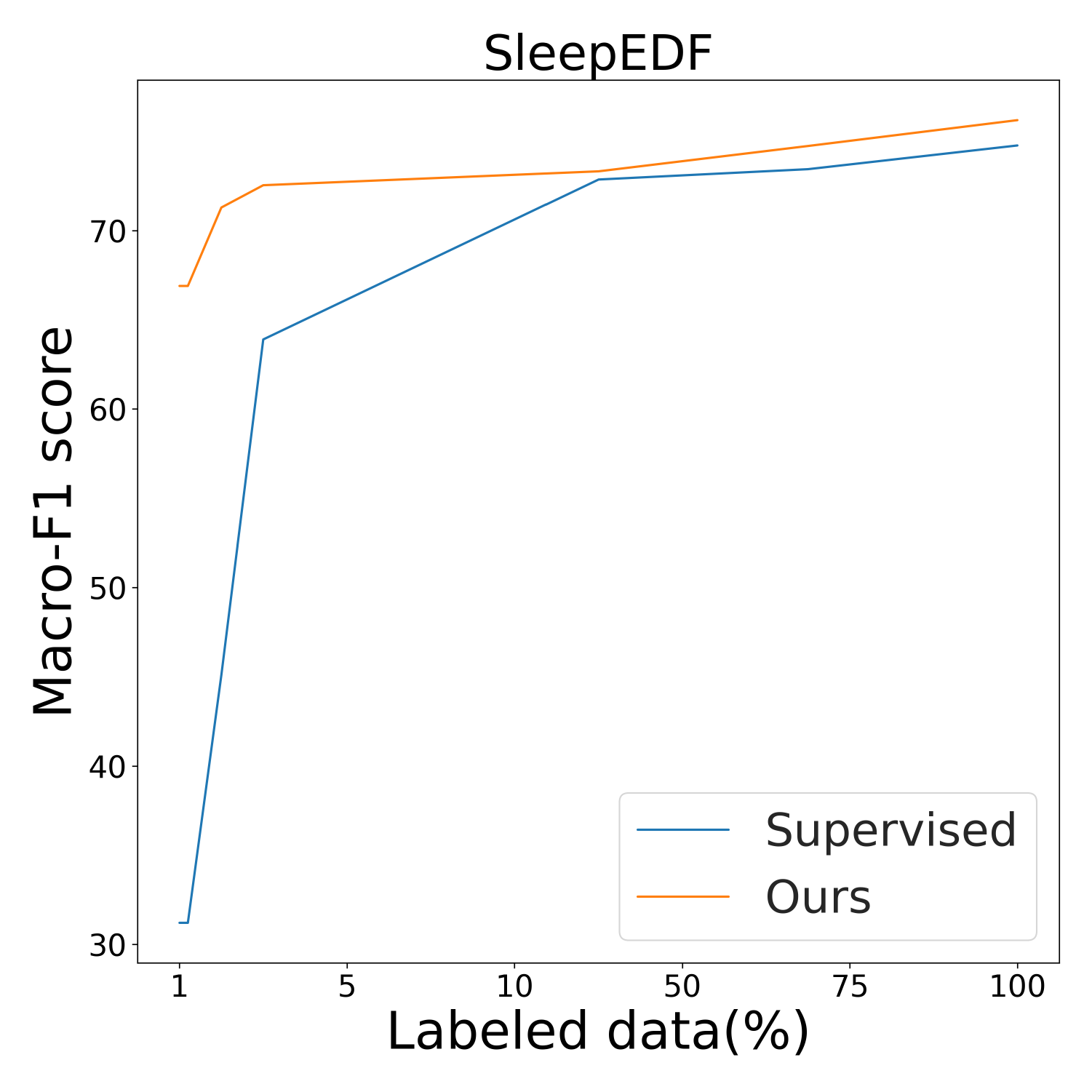} \\
\multicolumn{2}{c}{\includegraphics[width=41mm]{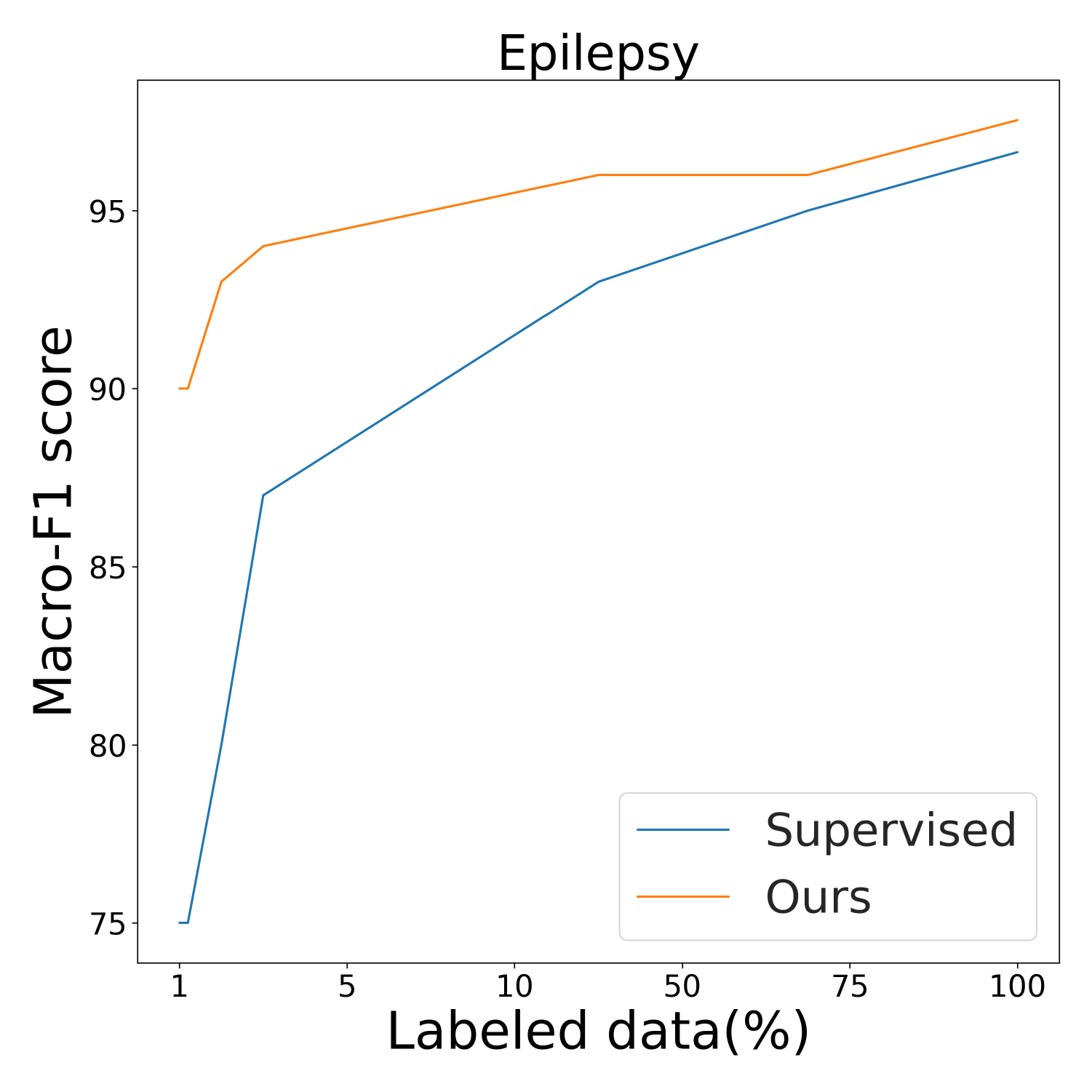} }\\

\end{tabular}
\caption{Comparison between supervised training vs. ours
fine-tuning for different few-labeled data scenarios in terms of Macro-F1 score. We observe from the plots that our method outperforms supervised training for all the datasets. It shows the quality of representation learned by the proposed model during unsupervised pre-training.}
\label{fig:sup}
\end{figure}

Our approach also shows its effectiveness in semi-supervised evaluation, in which we fine-tune the backbone CNN encoder along with the linear layer classifier on different subsets of the train data. Specifically, we created the subsets with varying percentages, including 1\%, 5\%, 10\%, 50\%, and 75\%. We presented these results in the form of plots for all the datasets in Fig.~\ref{fig:sup}, which provide a detailed comparison between the proposed SSL approach and traditional supervised training. For the subsets with a small percentage of labeled data (e.g., 1\%, 5\%, and 10\%), we observed a significant performance advantage for the proposed approach over the supervised training. This indicates that the representations learned during pretraining capture high-quality features from the unlabeled data, enabling the model to generalize effectively even with a limited number of labeled samples. This is particularly valuable in scenarios where acquiring labeled data is expensive or time-consuming. As the percentage of labeled data increases (e.g., 50\% and 75\%), the performance gap between the proposed SSL approach and supervised training gradually reduces. This suggests that the availability of more labeled samples allows the supervised training to catch up to some extent; however, it is important to note that the SSL approach still maintains a competitive advantage, consistently outperforming supervised training. This indicates that the pretraining phase successfully captures meaningful and discriminative features from the unlabeled data, enabling the model to leverage this knowledge and achieve superior performance compared to traditional supervised training. 

\begin{figure}[!ht]
\begin{minipage}[t]{0.24\textwidth}
    \includegraphics[height=3.3cm]{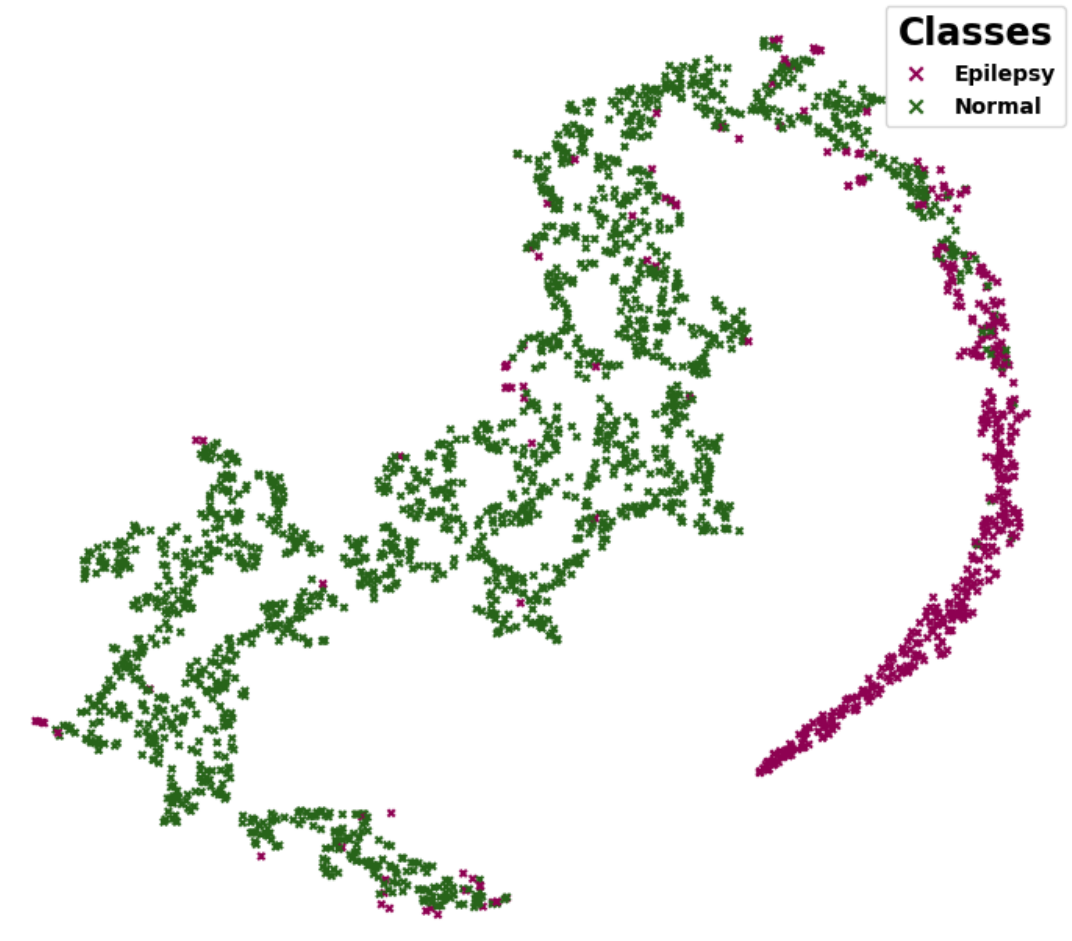}
    \includegraphics[height=3.3cm]{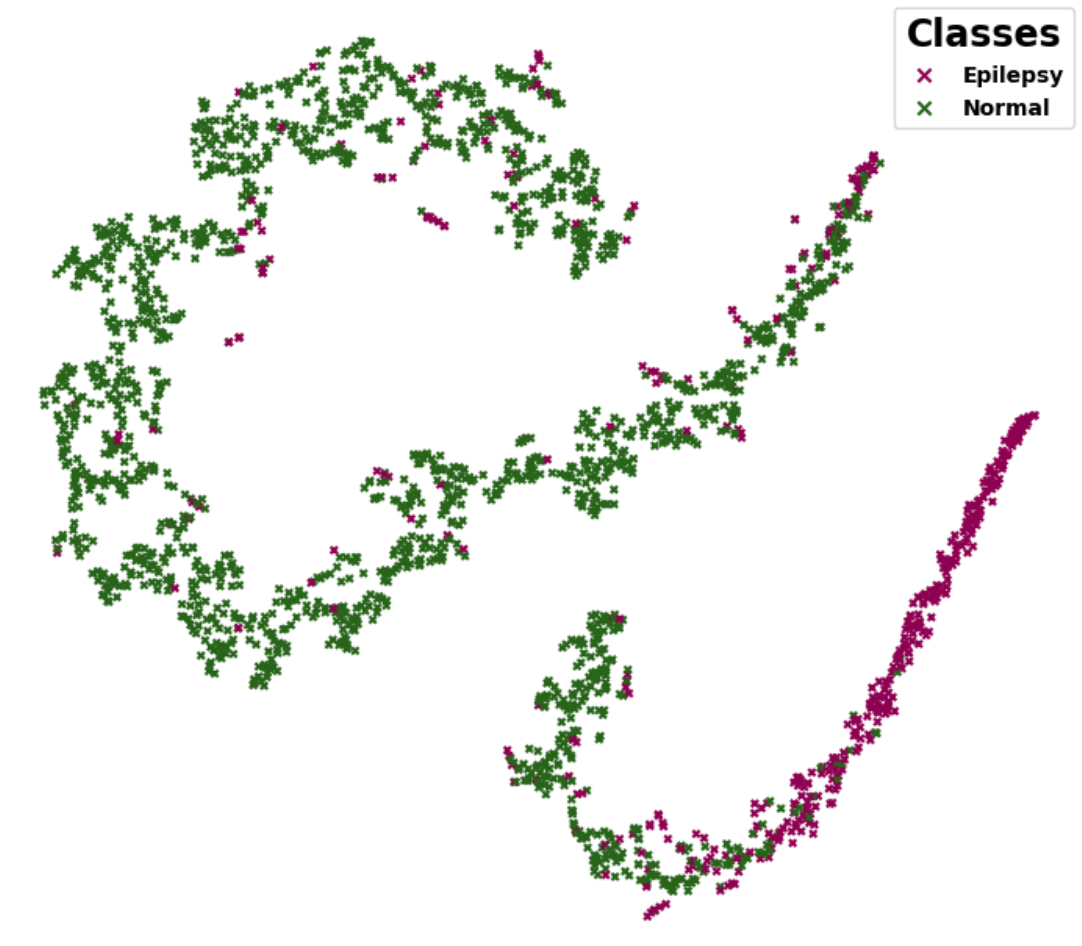}
\end{minipage} %
    \hfill%
\begin{minipage}[t]{0.24\textwidth}
    \includegraphics[height=3.3cm]{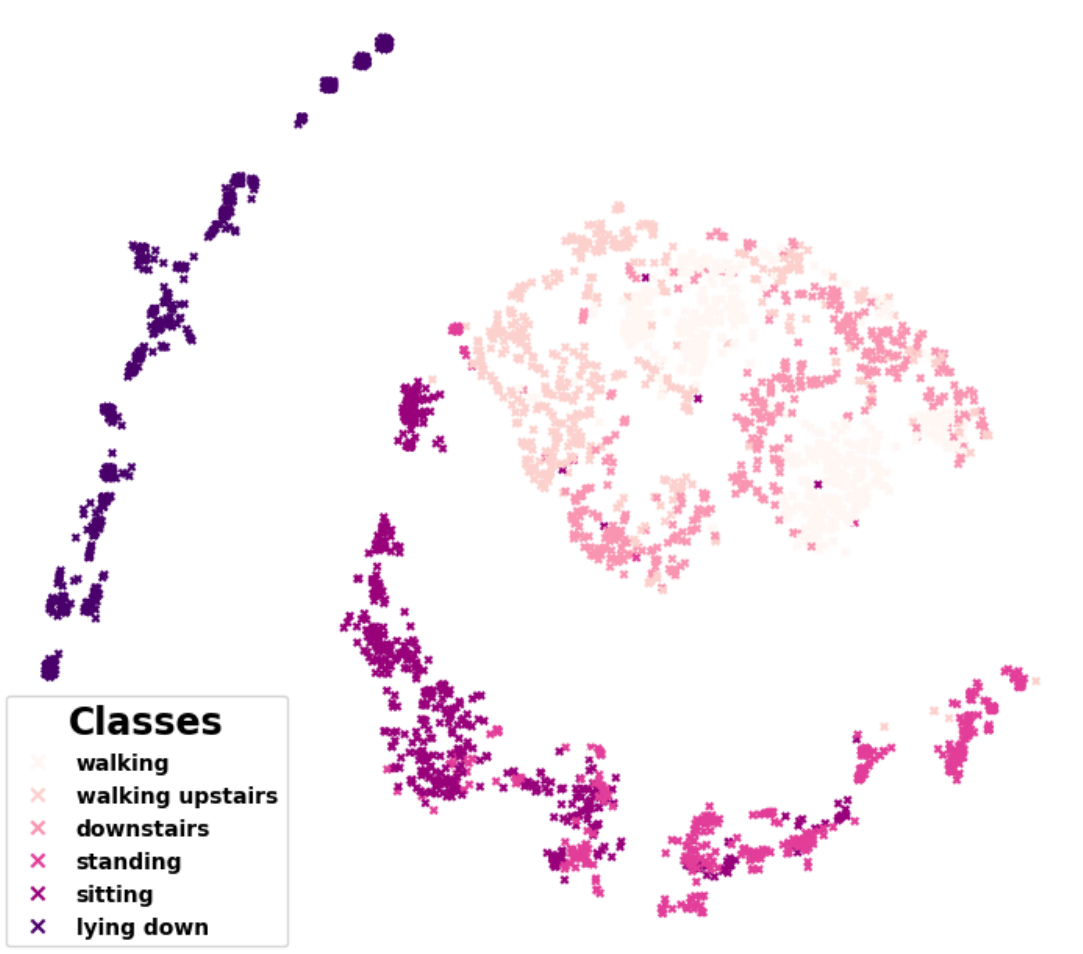}
    \includegraphics[height=3.3cm]{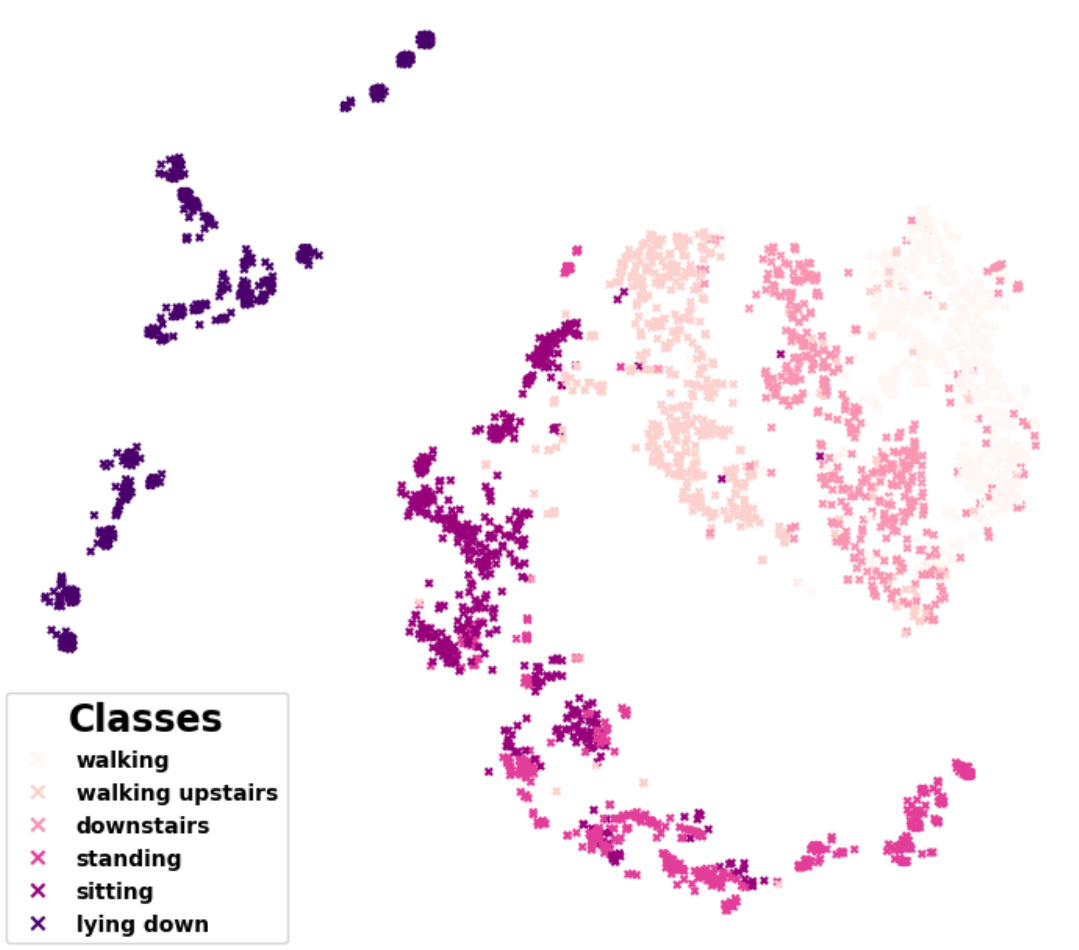}
\end{minipage}
\begin{minipage}{0.24\textwidth}
    \centering
    (a)
    \label{a}
\end{minipage}
\begin{minipage}{0.24\textwidth}
    \centering
    (b)
    \label{b}
\end{minipage}

\caption{t-SNE embedding visualization for supervised (first row) and ours (second row) on (a) Epilepsy and (b) HAR dataset, respectively. We observe a clear separation in the t-SNE plots corresponding to the proposed approach among different classes for both datasets. Also, the distribution of the class embeddings is better for the proposed SSL method than the supervised method.}
    \label{fig:tsne}
\end{figure}

Further, the t-SNE plots provide valuable insights into the quality of representations learned by the proposed approach compared to traditional supervised training. We specifically examined the t-SNE plots for the HAR and Epilepsy datasets (Fig.~\ref{fig:tsne}). These plots visualize how the different classes are distributed in the learned feature space. It is important to note that there is considerable overlap between the class embeddings in the t-SNE plots corresponding to the supervised training. In contrast, in the t-SNE plots corresponding to the representations learned by the proposed SSL approach, there is a clear and distinct separation between the representations of different classes for both the Epilepsy and HAR datasets. This indicates that the SSL approach has learned more discriminative features, allowing for better separation between classes, while the supervised method struggles to separate the classes effectively. Furthermore, when examining the distribution of class embeddings, we can see that the proposed method yields a more compact distribution compared to the supervised method. This suggests that the SSL approach has captured the relevant patterns and structures in the data more effectively compared to supervised training. 
These findings provide empirical evidence supporting the superiority of the proposed approach in learning representations that are more informative and discriminative for time-series classification tasks. This also demonstrates the effectiveness of leveraging self-supervised learning to enhance the quality of learned representations. 

We also consider t-SNE plots to compare the representations learned by our model and supervised baseline on HAR and Epilepsy dataset. The t-SNE plots provide valuable insights into the quality of representations learned by the proposed approach compared to traditional supervised training. We specifically examined the t-SNE plots for the HAR and Epilepsy datasets. The t-SNE plots present in Fig.~\ref{fig:tsne} (a) belongs to the Epilepsy dataset, whereas t-SNE plots in Fig.~\ref{fig:tsne} (b) belong to the HAR dataset, respectively. Plots in the first row of Fig.~\ref{fig:tsne} demonstrate the representations learned by the supervised method, while the plots in the second row demonstrate representations learned by the proposed approach. These plots visualize how the different classes are distributed in the learned feature space. It is important to note that there is considerable overlap between the class embeddings in the t-SNE plots corresponding to the supervised training. In contrast, in the t-SNE plots corresponding to the representations learned by the proposed SSL approach, there is a clear and distinct separation between the representations of different classes for both the Epilepsy and HAR datasets. This indicates that the SSL approach has learned more discriminative features, allowing for better separation between classes, while the supervised method struggles to separate the classes effectively. Furthermore, when examining the distribution of class embeddings, we can see that the proposed method yields a more compact distribution compared to the supervised method. This suggests that the SSL approach has captured the relevant patterns and structures in the data more effectively compared to supervised training. These findings provide empirical evidence supporting the superiority of the proposed approach in learning representations that are more informative and discriminative for time-series classification tasks. This also demonstrates the effectiveness of leveraging self-supervised learning to enhance the quality of learned representations.

\subsection{Computational Requirements} 
In the case of the proposed method, we aimed to strike a balance between performance and efficiency. It is worth noting that it offers advantages not only in terms of performance but also in computational efficiency compared to some of the baselines, specifically contrastive methods like CPC~\cite{cpc}, SimCLR~\cite{simclr}, MoCo~\cite{moco} and TS-TCC~\cite{tstcc}. For example, TS-TCC is primarily uses multi-task contrastive loss and transformers as the backbone encoder, contributing to increased computational resources and time needed for training and inference. Similarly, SimCLR also often requires substantial computational resources due to the use of large negative sample sets and the need for extensive training iterations to learn desired representations. Similarly, CPC is also computationally demanding as it uses an autoregressive model to generate latent codes, which are then contrasted against a set of negative samples to learn informative representations. The autoregressive model used in CPC requires sequential processing, which limits parallel processing and increases training time. Additionally, generating negative samples for contrastive learning also increases memory usage and computational complexity. On the other hand, the proposed approach utilizes TCN architecture which allows for parallel and efficient processing of temporal data due to its ability to capture long-range dependencies. Additionally, the proposed approach follows the non-contrastive SSL paradigm, which results in a simplified training pipeline as it eliminates the need for negative sampling. However, when comparing to other non-contrastive methods like SimSiam~\cite{simsiam}, our approach incurs slightly higher parameter count due to the additional TCN head in both online and target network branches. This TCN head driven parameter increase is reasonable considering the performance gains over SimSiam.
These benefits contribute to the effectiveness of our approach for learning high-quality representations from time-series data. The proposed method is developed using PyTorch version 1.10 and is compatible with Python version 3.7 or higher. The experiments are conducted on a machine equipped with an NVIDIA Tesla V100 GPU.

\subsection{Ablation studies}
To show the significance of each module in our model, we perform an extensive ablation study. The ablation study is divided into four parts: (1) the effect of head Branches; (2) the effect of TCN Position (3) the effect of dilation and kernel Size; (4) the effect of augmentations.

\subsubsection{Effect of Head Branches}
To understand how much TCN and MLP branch is contributing to the performance of our model, we train our model by \textbf{Case(1):} Removing MLP head branches from both online and target network, \textbf{Case(2):} Removing TCN head branches from both online and target network. 
In both cases, we observe lower accuracy compared to our main results across all the datasets.
Note that in case(2), the proposed model becomes equivalent to BYOl~\cite{byol}, which does not have TCN heads. All results are shown in Table~\ref{tab:table2} and~\ref{tab:table3}.
HAR, IMU sensor, and ECG MEDH dataset are sampled at 50Hz sampling frequency and are dominated by low-frequency components. On the other hand, SleepEDF and Epilepsy datasets are sampled at 100Hz and 173.5Hz, respectively. These contain an adequate amount of high-frequency components. As can be seen in Table~\ref{tab:table2} and~\ref{tab:table3}, a combination of TCN and MLP provides improvement over TCN or MLP alone for these datasets. Overall this shows that the backbone is efficiently capturing global features, and TCN then hierarchically captures features at various levels in the global feature space. Combining both modules further boosts the performance to outperform the current state-of-the-art method TS-TCC.

\subsubsection{Effect of TCN Position}
To validate the efficiency of the convolution network as the backbone encoders, we did experiments with TCN and MLP in the backbone network. 
As mentioned above in section 3, TCN  captures high-frequency components while MLP captures low-frequency components from the time series data. 
In this experiment, we pass raw input directly into the TCN and MLP head in both the online and the target branches to extract high- and low-frequency components from the input signal. The  extracted high- and low-frequency components are then processed by the CNN encoders in each branch of the network. The embeddings learned by the CNN encoders have both high- and low-frequency components, which are utilized further by the projection and prediction heads of each branch. The projection and prediction heads are similar to our main experiments and follow the same processing pipeline as described in the methodology section.

The last row of Table~\ref{tab:table2} and~\ref{tab:table3} corresponds to the numbers obtained from this experiment. 
The accuracy and MF1 obtained under this setting are comparable to our main results. However, this setup results in a huge increment in the computational overhead in terms of space and time complexity. 
Besides the main architecture, the network has extra TCN and MLP heads in both the online and target branch of the network, which essentially increases the training time and also the memory footprints of the GPU. This experiment validates the effectiveness of our proposed architecture which outperforms all the baselines with comparatively lower computational requirements.  

\subsubsection{Effect of Augmentation}
Data augmentation is an important part of self-supervised learning. We train our model on the HAR dataset by applying. (1) Same augmentation: augmentation from the same family. We apply Jitter-Permutation-Rotation on both sides. Rotation is \(30^\circ\) and \(45^\circ\)  for each augmentation, respectively; (2) Different augmentation: augmentation from different families. We applied Jitter-Permutation-Rotation as a strong augmentation and jitter-scale as a weak augmentation. As shown in Table~\ref{tab:aug}, the same augmentations result in better learning of the network.

\begin{table}[ht]
    \centering
    \begin{adjustbox}{width=0.3\textwidth}
    \begin{tabular}{cc}
    \toprule
    \multirow{1}{*}{Methods} & \multicolumn{1}{c}{HAR} \\
    \midrule
    Same Aug & 93.28 $\pm$ 0.30\% \\
    Different Aug & 78.78 $\pm$ 1.25\% \\      
    \bottomrule
    \end{tabular}
    \end{adjustbox}
    \caption{Performance of our model on under different augmentations. Metrics is Macro-F1 score.}
    \label{tab:aug}
\end{table}

\subsubsection{Effect of Kernel and Dilation}
Choosing the correct size for kernel and dilation is an important factor in our model. We train our model for different kernel and dilation sizes. As discussed above in the proposed method the kernel size of TCN must be larger than the kernel size of the backbone encoder and the dilation rate of TCN must be greater than the kernel size of TCN. We prove this empirically by training our model in two different settings $(K1<K2, D>K1)-K2$ and $(K1>K2, D<K1)-K2$, where $K1$ and $D$ is a kernel of TCN and $K2$, is the kernel of backbone. Results are shown in Table ~\ref{tab:kd}.
\begin{table}[ht]
    \centering
    \begin{adjustbox}{width=0.35\textwidth}
    \begin{tabular}{cc}
        \toprule
         \multirow{1}{*}{Methods} & \multicolumn{1}{c}{HAR} \\
        \midrule
        $(K1<K2,D>K1)-K2$ & 93.28 $\pm$ 0.36 \\
        $(K1>K2,D<K1)-K2$ & 89.90 $\pm$ 1.26 \\
        
        \bottomrule
    \end{tabular}
    \end{adjustbox}
    \caption{Performance of our model under different Dilation and Kernel rates. Metrics are Macro-F1 score.}
    \label{tab:kd}
\end{table}

\subsubsection{Sensitivity of Balance Factor in Loss}
We performed a sensitivity analysis of \(\lambda\) on the HAR dataset. \(\lambda\) is a balance factor in our loss function. Selecting the wrong lambda values leads to lower performance. We train our model for four different combinations of \(\lambda\) values. Results are shown in Table ~\ref{tab:bal_f}. It shows that very large or small values of $\lambda$ result in suboptimal performance.
\begin{table}[ht]
    \centering
    \begin{adjustbox}{width=0.25\textwidth}
    \begin{tabular}{cc}
        \toprule
         \multirow{1}{*}{Methods} & \multicolumn{1}{c}{HAR} \\
        \midrule
        \(\lambda=0.005\) & 91.78 $\pm$ 0.31 \\
        \(\lambda=0.5\) & 93.28 $\pm$ 0.26 \\
        
        \(\lambda=5\) & 88.90 $\pm$ 1.26 \\
        
        \(\lambda=500\) & 87.56 $\pm$ 13.26 \\
        
        \bottomrule
    \end{tabular}
    \end{adjustbox}
    \caption{Performance of our model on under different $\lambda$ values for HAR dataset. Metrics are Macro-F1 score.}
    \label{tab:bal_f}
\end{table}

Following the ablation studies, we find that, using augmentations from the same augmentation family results in better performance of the model. Further, we also find that keeping the kernel size of TCN larger than the kernel size of the backbone encoder and the dilation rate of TCN greater than the kernel size of TCN generally gives the best results. Also, assigning equal weight($\lambda$) to both low- and high-frequency modules (LFFB and HFFB) results in the best performance.

\section{Limitations}
The proposed method for learning representations from unlabeled time series data presents advantages over existing approaches; however, it is also important to highlight the limitations and potential areas for improvement. One limitation is the specific set of augmentations used in the proposed method, namely the jitter-permute-rotate augmentations. While these augmentations are effective for capturing certain types of variations in the time series data, they may not be suitable for all types of datasets or domains. Investigating a broader range of augmentations, tailored toward time series data could enhance the generalization ability of the learned representations. 
Additionally, the proposed method is evaluated on five real-world time series datasets, demonstrating superior performance. However, the generalizability of the method to other domains remains unexplored. Conducting experiments on a wider range of diverse datasets, including multimodal time series data and applications, would provide a more comprehensive assessment of the method's performance and robustness. Furthermore, the proposed method has limited interpretability of the learned parameters and intermediate representations. While the model achieves promising performance on various time-series datasets, understanding the underlying reasons for its decisions may require additional effort. Future work could focus on incorporating interpretability techniques into the proposed approach.

\section{Conclusion}
In this paper, we propose a novel self-supervised learning approach for time-series representation learning. Unlike existing contrastive approaches, our proposed approach does not use negative pairs to learn representation. First, it creates two different augmentations and passes them into the encoder to learn the desired representation through HFFB and LFFB modules. LHFFB is further deployed to learn complementary representation which LFFB and HFFB are unable to learn individually. In this way, the model learns robust representation from raw time series data with very few labels. By using only 10\% of the labeled data, the proposed method can achieve close performance to the supervised training with full labeled data. The proposed approach has advantages but also has certain limitations. The specific set of augmentations used in the method may not be suitable for all datasets and domains, warranting further exploration of tailored augmentations for time series data. Additionally, the generalizability of the method to different datasets and domains needs to be investigated through experiments on diverse datasets. The limited interpretability of the learned representations is another limitation, which could be addressed by incorporating interpretability techniques into the approach.
Another important aspect to consider in time series analysis is the presence of multi-modality, so it may be beneficial to explore alternative SSL architectures in the future that can better capture the multi-modality present in time series data. The proposed idea would provide motivation and future direction to develop more SSL approaches for time series data, which take into account the nature of the problem.

\bibliographystyle{IEEEtran}
\bibliography{tnnls}

\begin{thebibliography}{10}
\providecommand{\url}[1]{#1}
\csname url@samestyle\endcsname
\providecommand{\newblock}{\relax}
\providecommand{\bibinfo}[2]{#2}
\providecommand{\BIBentrySTDinterwordspacing}{\spaceskip=0pt\relax}
\providecommand{\BIBentryALTinterwordstretchfactor}{4}
\providecommand{\BIBentryALTinterwordspacing}{\spaceskip=\fontdimen2\font plus
\BIBentryALTinterwordstretchfactor\fontdimen3\font minus \fontdimen4\font\relax}
\providecommand{\BIBforeignlanguage}[2]{{%
\expandafter\ifx\csname l@#1\endcsname\relax
\typeout{** WARNING: IEEEtran.bst: No hyphenation pattern has been}%
\typeout{** loaded for the language `#1'. Using the pattern for}%
\typeout{** the default language instead.}%
\else
\language=\csname l@#1\endcsname
\fi
#2}}
\providecommand{\BIBdecl}{\relax}
\BIBdecl

\bibitem{faust2018deep}
O.~Faust, Y.~Hagiwara, T.~J. Hong, O.~S. Lih, and U.~R. Acharya, ``Deep learning for healthcare applications based on physiological signals: A review,'' \emph{Computer methods and programs in biomedicine}, vol. 161, pp. 1--13, 2018.

\bibitem{8614252}
S.~Siami-Namini, N.~Tavakoli, and A.~Siami~Namin, ``A comparison of arima and lstm in forecasting time series,'' in \emph{2018 17th IEEE International Conference on Machine Learning and Applications (ICMLA)}, 2018, pp. 1394--1401.

\bibitem{8437249}
C.-L. Liu, W.-H. Hsaio, and Y.-C. Tu, ``Time series classification with multivariate convolutional neural network,'' \emph{IEEE Transactions on Industrial Electronics}, vol.~66, no.~6, pp. 4788--4797, 2019.

\bibitem{zhang2022tn}
L.~Zhang, X.~Chang, J.~Liu, M.~Luo, Z.~Li, L.~Yao, and A.~Hauptmann, ``Tn-zstad: Transferable network for zero-shot temporal activity detection,'' \emph{IEEE Transactions on Pattern Analysis and Machine Intelligence}, vol.~45, no.~3, pp. 3848--3861, 2022.

\bibitem{Ching142760}
T.~Ching, D.~S. Himmelstein, B.~K. Beaulieu-Jones, A.~A. Kalinin, B.~T. Do, G.~P. Way, E.~Ferrero, P.-M. Agapow, M.~Zietz, M.~M. Hoffman \emph{et~al.}, ``Opportunities and obstacles for deep learning in biology and medicine,'' \emph{Journal of The Royal Society Interface}, vol.~15, no. 141, p. 20170387, 2018.

\bibitem{chang2021comprehensive}
X.~Chang, P.~Ren, P.~Xu, Z.~Li, X.~Chen, and A.~Hauptmann, ``A comprehensive survey of scene graphs: Generation and application,'' \emph{IEEE Transactions on Pattern Analysis and Machine Intelligence}, vol.~45, no.~1, pp. 1--26, 2021.

\bibitem{oord2018representation}
A.~v.~d. Oord, Y.~Li, and O.~Vinyals, ``Representation learning with contrastive predictive coding,'' \emph{arXiv preprint arXiv:1807.03748}, 2018.

\bibitem{moco}
K.~He, H.~Fan, Y.~Wu, S.~Xie, and R.~Girshick, ``Momentum contrast for unsupervised visual representation learning,'' 2020.

\bibitem{simclr}
T.~Chen, S.~Kornblith, M.~Norouzi, and G.~Hinton, ``A simple framework for contrastive learning of visual representations,'' 2020.

\bibitem{li2022video}
M.~Li, P.-Y. Huang, X.~Chang, J.~Hu, Y.~Yang, and A.~Hauptmann, ``Video pivoting unsupervised multi-modal machine translation,'' \emph{IEEE Transactions on Pattern Analysis and Machine Intelligence}, vol.~45, no.~3, pp. 3918--3932, 2022.

\bibitem{jigsaw}
M.~Noroozi and P.~Favaro, ``Unsupervised learning of visual representations by solving jigsaw puzzles,'' 2017.

\bibitem{exemplecnn}
A.~Dosovitskiy, P.~Fischer, J.~T. Springenberg, M.~Riedmiller, and T.~Brox, ``Discriminative unsupervised feature learning with exemplar convolutional neural networks,'' 2015.

\bibitem{rotpred}
S.~Gidaris, P.~Singh, and N.~Komodakis, ``Unsupervised representation learning by predicting image rotations,'' 2018.

\bibitem{colorization}
R.~Zhang, P.~Isola, and A.~A. Efros, ``Colorful image colorization,'' 2016.

\bibitem{cpc}
A.~van~den Oord, Y.~Li, and O.~Vinyals, ``Representation learning with contrastive predictive coding,'' 2019.

\bibitem{contextautoenc}
D.~Pathak, P.~Krahenbuhl, J.~Donahue, T.~Darrell, and A.~A. Efros, ``Context encoders: Feature learning by inpainting,'' 2016.

\bibitem{brainsplit}
R.~Zhang, P.~Isola, and A.~A. Efros, ``Split-brain autoencoders: Unsupervised learning by cross-channel prediction,'' 2017.

\bibitem{liu2021self}
X.~Liu, F.~Zhang, Z.~Hou, L.~Mian, Z.~Wang, J.~Zhang, and J.~Tang, ``Self-supervised learning: Generative or contrastive,'' \emph{IEEE Transactions on Knowledge and Data Engineering}, 2021.

\bibitem{byol}
J.-B. Grill, F.~Strub, F.~Altché, C.~Tallec, P.~H. Richemond, E.~Buchatskaya, C.~Doersch, B.~A. Pires, Z.~D. Guo, M.~G. Azar, B.~Piot, K.~Kavukcuoglu, R.~Munos, and M.~Valko, ``Bootstrap your own latent: A new approach to self-supervised learning,'' 2020.

\bibitem{simsiam}
X.~Chen and K.~He, ``Exploring simple siamese representation learning,'' 2020.

\bibitem{sslecg}
\BIBentryALTinterwordspacing
P.~Sarkar and A.~Etemad, ``Self-supervised ecg representation learning for emotion recognition,'' \emph{IEEE Transactions on Affective Computing}, p. 1–1, 2021. [Online]. Available: \url{http://dx.doi.org/10.1109/TAFFC.2020.3014842}
\BIBentrySTDinterwordspacing

\bibitem{selfhar}
\BIBentryALTinterwordspacing
C.~I. Tang, I.~Perez-Pozuelo, D.~Spathis, S.~Brage, N.~Wareham, and C.~Mascolo, ``Selfhar,'' \emph{Proceedings of the ACM on Interactive, Mobile, Wearable and Ubiquitous Technologies}, vol.~5, no.~1, p. 1–30, Mar 2021. [Online]. Available: \url{http://dx.doi.org/10.1145/3448112}
\BIBentrySTDinterwordspacing

\bibitem{seqclr}
A.~Aberdam, R.~Litman, S.~Tsiper, O.~Anschel, R.~Slossberg, S.~Mazor, R.~Manmatha, and P.~Perona, ``Sequence-to-sequence contrastive learning for text recognition,'' 2020.

\bibitem{franceschi2020unsupervised}
J.-Y. Franceschi, A.~Dieuleveut, and M.~Jaggi, ``Unsupervised scalable representation learning for multivariate time series,'' 2020.

\bibitem{iwana2021empirical}
B.~K. Iwana and S.~Uchida, ``An empirical survey of data augmentation for time series classification with neural networks,'' \emph{Plos one}, vol.~16, no.~7, p. e0254841, 2021.

\bibitem{wavelets}
\BIBentryALTinterwordspacing
A.~Graps, ``An introduction to wavelets,'' \emph{IEEE Comput. Sci. Eng.}, vol.~2, no.~2, p. 50–61, jun 1995. [Online]. Available: \url{https://doi.org/10.1109/99.388960}
\BIBentrySTDinterwordspacing

\bibitem{tcn}
C.~Lea, M.~D. Flynn, R.~Vidal, A.~Reiter, and G.~D. Hager, ``Temporal convolutional networks for action segmentation and detection,'' 2016.

\bibitem{RL}
Y.~Bengio, A.~Courville, and P.~Vincent, ``Representation learning: A review and new perspectives,'' 2014.

\bibitem{kingma2013auto}
D.~P. Kingma and M.~Welling, ``Auto-encoding variational bayes,'' \emph{arXiv preprint arXiv:1312.6114}, 2013.

\bibitem{ashfahani2020devdan}
A.~Ashfahani, M.~Pratama, E.~Lughofer, and Y.-S. Ong, ``Devdan: Deep evolving denoising autoencoder,'' \emph{Neurocomputing}, vol. 390, pp. 297--314, 2020.

\bibitem{miotto2016deep}
R.~Miotto, L.~Li, B.~A. Kidd, and J.~T. Dudley, ``Deep patient: an unsupervised representation to predict the future of patients from the electronic health records,'' \emph{Scientific reports}, vol.~6, no.~1, pp. 1--10, 2016.

\bibitem{deng2013recent}
L.~Deng, J.~Li, J.-T. Huang, K.~Yao, D.~Yu, F.~Seide, M.~Seltzer, G.~Zweig, X.~He, J.~Williams \emph{et~al.}, ``Recent advances in deep learning for speech research at microsoft,'' in \emph{2013 IEEE international conference on acoustics, speech and signal processing}.\hskip 1em plus 0.5em minus 0.4em\relax IEEE, 2013, pp. 8604--8608.

\bibitem{shewalkar2019performance}
A.~Shewalkar, ``Performance evaluation of deep neural networks applied to speech recognition: Rnn, lstm and gru,'' \emph{Journal of Artificial Intelligence and Soft Computing Research}, vol.~9, no.~4, pp. 235--245, 2019.

\bibitem{yan2021zeronas}
C.~Yan, X.~Chang, Z.~Li, W.~Guan, Z.~Ge, L.~Zhu, and Q.~Zheng, ``Zeronas: Differentiable generative adversarial networks search for zero-shot learning,'' \emph{IEEE transactions on pattern analysis and machine intelligence}, vol.~44, no.~12, pp. 9733--9740, 2021.

\bibitem{choi2016multi}
E.~Choi, M.~T. Bahadori, E.~Searles, C.~Coffey, M.~Thompson, J.~Bost, J.~Tejedor-Sojo, and J.~Sun, ``Multi-layer representation learning for medical concepts,'' in \emph{proceedings of the 22nd ACM SIGKDD international conference on knowledge discovery and data mining}, 2016, pp. 1495--1504.

\bibitem{yuan2019wave2vec}
Y.~Yuan, G.~Xun, Q.~Suo, K.~Jia, and A.~Zhang, ``Wave2vec: Deep representation learning for clinical temporal data,'' \emph{Neurocomputing}, vol. 324, pp. 31--42, 2019.

\bibitem{mikolov2013distributed}
T.~Mikolov, I.~Sutskever, K.~Chen, G.~S. Corrado, and J.~Dean, ``Distributed representations of words and phrases and their compositionality,'' \emph{Advances in neural information processing systems}, vol.~26, 2013.

\bibitem{zbontar2021barlow}
J.~Zbontar, L.~Jing, I.~Misra, Y.~LeCun, and S.~Deny, ``Barlow twins: Self-supervised learning via redundancy reduction,'' in \emph{International Conference on Machine Learning}.\hskip 1em plus 0.5em minus 0.4em\relax PMLR, 2021, pp. 12\,310--12\,320.

\bibitem{tstcc}
E.~Eldele, M.~Ragab, Z.~Chen, M.~Wu, C.~K. Kwoh, X.~Li, and C.~Guan, ``Time-series representation learning via temporal and contextual contrasting,'' 2021.

\bibitem{bai2018empirical}
S.~Bai, J.~Z. Kolter, and V.~Koltun, ``An empirical evaluation of generic convolutional and recurrent networks for sequence modeling,'' \emph{arXiv preprint arXiv:1803.01271}, 2018.

\bibitem{zhang2021temporal}
H.~Zhang, W.~Hu, D.~Cao, Q.~Huang, Z.~Chen, and F.~Blaabjerg, ``A temporal convolutional network based hybrid model of short-term electricity price forecasting,'' \emph{CSEE Journal of Power and Energy Systems}, 2021.

\bibitem{yan2020temporal}
J.~Yan, L.~Mu, L.~Wang, R.~Ranjan, and A.~Y. Zomaya, ``Temporal convolutional networks for the advance prediction of enso,'' \emph{Scientific reports}, vol.~10, no.~1, pp. 1--15, 2020.

\bibitem{dai2022price}
W.~Dai, Y.~An, and W.~Long, ``Price change prediction of ultra high frequency financial data based on temporal convolutional network,'' \emph{Procedia Computer Science}, vol. 199, pp. 1177--1183, 2022.

\bibitem{sleepstage}
\BIBentryALTinterwordspacing
A.~B. Colten~HR, ``Stages of sleep,'' 2006. [Online]. Available: \url{https://www.ncbi.nlm.nih.gov/books/NBK19956/}
\BIBentrySTDinterwordspacing

\bibitem{anguita2013public}
D.~Anguita, A.~Ghio, L.~Oneto, X.~Parra, J.~L. Reyes-Ortiz \emph{et~al.}, ``A public domain dataset for human activity recognition using smartphones.'' in \emph{Esann}, vol.~3, 2013, p.~3.

\bibitem{goldberger2000physiobank}
A.~L. Goldberger, L.~A. Amaral, L.~Glass, J.~M. Hausdorff, P.~C. Ivanov, R.~G. Mark, J.~E. Mietus, G.~B. Moody, C.-K. Peng, and H.~E. Stanley, ``Physiobank, physiotoolkit, and physionet: components of a new research resource for complex physiologic signals,'' \emph{circulation}, vol. 101, no.~23, pp. e215--e220, 2000.

\bibitem{andrzejak2001indications}
R.~G. Andrzejak, K.~Lehnertz, F.~Mormann, C.~Rieke, P.~David, and C.~E. Elger, ``Indications of nonlinear deterministic and finite-dimensional structures in time series of brain electrical activity: Dependence on recording region and brain state,'' \emph{Physical Review E}, vol.~64, no.~6, p. 061907, 2001.

\bibitem{IMU}
F.~Lomio, E.~Skenderi, D.~Mohamadi, J.~Collin, R.~Ghabcheloo, and H.~Huttunen, ``Surface type classification for autonomous robot indoor navigation,'' \emph{arXiv preprint arXiv:1905.00252}, 2019.

\bibitem{Dua:2019}
\BIBentryALTinterwordspacing
D.~Dua and C.~Graff, ``{UCI} machine learning repository,'' 2017. [Online]. Available: \url{http://archive.ics.uci.edu/dataset/319/mhealth+dataset}
\BIBentrySTDinterwordspacing

\end{thebibliography}

\end{document}